\documentclass[sigconf]{acmart}

\acmYear{2026}\copyrightyear{2026}
\setcopyright{cc}
\setcctype[4.0]{by}
\acmConference[ACM CAIS '26]{ACM Conference on AI and Agentic Systems}{May 26--29, 2026}{San Jose, CA, USA}
\acmBooktitle{ACM Conference on AI and Agentic Systems (ACM CAIS '26), May 26--29, 2026, San Jose, CA, USA}
\acmDOI{10.1145/3786335.3813147}
\acmISBN{979-8-4007-2415-2/26/05}

\usepackage{tabularx}
\usepackage{booktabs}
\usepackage{hyperref}
\usepackage{multirow}
\usepackage{listings}

\usepackage{makecell}
\usepackage{graphicx}
\usepackage{subcaption}
\usepackage{wrapfig}
\usepackage{framed}
\usepackage{amsthm}
\usepackage{xcolor}
\usepackage{flushend}
\usepackage{pifont} 
\usepackage{amsmath}
\usepackage{paralist}
\usepackage{siunitx}
\usepackage{enumitem}
\usepackage{algorithm}
\usepackage[noend]{algorithmic}
\usepackage{fancyhdr}
\usepackage{xcolor}
\usepackage{xspace}
\usepackage{ragged2e}

\newcolumntype{L}[1]{>{\RaggedRight\arraybackslash}p{#1}}

\newcommand{\cut}[1]{}

\newcommand{\Space}[1]{}

\colorlet{shadecolor}{gray!20}
\definecolor{Gray}{gray}{0.8}

\newtheoremstyle{findingstyle}
  {0pt}   
  {0pt}   
  {\itshape}  
  {0pt}       
  {\bfseries} 
  {.}         
  {5pt plus 1pt minus 1pt} 
  {}          

\theoremstyle{findingstyle}
\newtheorem{findinner}{\textbf{Finding}}

\newenvironment{find}
  {\begin{shaded}\begin{findinner}}
  {\end{findinner}\end{shaded}}

\newcommand{\finding}[1]{
  \begin{find}
    #1
  \end{find}
}

\newcommand{\numTasks}{32\xspace}      
\newcommand{\numRuns}{614\xspace}      
\newcommand{\numLLMs}{3\xspace}        

\newcommand{\pctSilentCorruption}{15.3\%\xspace}     
\newcommand{\pctBehavioralDetours}{40.3\%\xspace}    
\newcommand{\pctRerouting}{80.6\%\xspace}            
\newcommand{\pctExtendedExecution}{37.4\%\xspace}    
\newcommand{\pctEarlyTermination}{25.3\%\xspace}     

\newcommand{\pctLowCostIncorrect}{76.2\%\xspace}     
\newcommand{\pctHighCostCorrect}{16.3\%\xspace}      

\newcommand{\costBehavioralDetours}{1.5$\times$\xspace}  
\newcommand{\costExtendedExecution}{2.1$\times$\xspace}   
\begin{document}

\title{Trace-Level Analysis of Information Contamination in Multi-Agent Systems}

\author{Anna Mazhar}
\affiliation{\institution{Cornell University}\country{Ithaca, NY, USA}}
\author{Huzaifa Suri}
\affiliation{\institution{University of Illinois}\country{Urbana-Champaign, IL, USA}}
\author{Sainyam Galhotra}
\affiliation{\institution{Cornell University}\country{Ithaca, NY, USA}}

\begin{abstract}
Reasoning over heterogeneous artifacts (PDFs, spreadsheets, slide decks, etc.) 
increasingly occurs within structured agent workflows that iteratively extract, 
transform, and reference external information. In these workflows, 
uncertainty is not merely an input-quality issue: 
it can redirect decomposition and routing decisions, reshape intermediate state, 
and produce qualitatively different execution trajectories. 
We study this phenomenon by treating uncertainty as a controlled variable:
we inject structured perturbations into artifact-derived representations, 
execute fixed workflows under comprehensive logging, and 
quantify contamination via trace divergence in plans, tool invocations, and intermediate state. 
Across \numRuns paired runs on \numTasks GAIA tasks with three different language models, we find a decoupling: 
workflows may diverge substantially yet recover correct answers, or remain structurally similar 
while producing incorrect outputs. We characterize three manifestation types: silent semantic 
corruption, behavioral detours with recovery, and combined structural disruption and their 
control-flow signatures (rerouting, extended execution, early termination). We measure operational 
costs and characterize why commonly used verification guardrails fail to intercept contamination. 
We contribute (i)~a formal taxonomy of contamination manifestations in structured workflows, 
(ii)~a trace-based measurement framework for detecting and localizing contamination across agent 
interactions, and (iii)~empirical evidence with implications for targeted verification, defensive 
design, and cost control.
\end{abstract}
\maketitle

\vspace{-6pt}
\section{Introduction}

AI agents increasingly operate over heterogeneous external artifacts such as PDF reports, spreadsheets, slide decks, and semi-structured documents,
whose contents must be extracted, normalized, and referenced across multiple reasoning steps~\cite{yao2023react,toolformer2023}.
In such workflows, information extracted from external sources becomes embedded in intermediate state
and directly influences task decomposition, tool invocation, and coordination among agents.
Consequently, errors introduced during extraction do not remain localized; they shape subsequent reasoning steps and system behavior~\cite{kim2025sciencescalingagentsystems}.

This challenge is particularly pronounced in structured multi-agent systems~\cite{wu2024autogen,hong2024metagpt},
where specialization by role, tool access, and planning responsibility introduces explicit information-exchange boundaries.
Data extracted by one component are interpreted, transformed, and reused by others.
While this modular design improves scalability and separation of concerns,
it also creates new pathways for error propagation~\cite{cemri2025multiagentllmsystemsfail}.
A key failure mode is data that is locally valid but globally corrupting: corrupted extractions, truncated tool outputs, or misaligned table schemas~\cite{mialon2024gaia} 
that satisfy local syntactic checks while distorting downstream computation. 
Since intermediate results often appear well-formed, failures emerge indirectly, through unexpected behavior, increased workflow complexity, or inconsistent outputs.
 
Despite this structural vulnerability, prevailing evaluation practices focus primarily on endpoint accuracy:
does the final output match a reference answer? Such evaluation collapses the internal dynamics of
the workflow, providing limited insight into how uncertainty propagates, under what conditions it amplifies,
and where validation mechanisms should intervene. From an information systems perspective, 
this leaves critical design questions unanswered regarding interface contracts, 
invariant enforcement, and cross-boundary verification.


\begin{example} \label{para:example_failure_mode}
Consider a workflow tasked with analyzing quarterly financial data to answer: ``Which division had the highest revenue growth?''
A table parser misidentifies merged header cells, causing downstream queries to reference incorrect columns.
A data analysis agent computes growth rates that exceed total company revenue.
Rather than failing immediately, the planner proposes alternative interpretations,
retries extraction with modified parser settings,
invokes cross-validation routines,
and explores competing table-structure hypotheses.
The execution expands from three to nine steps.
A simple schema-level invariant that verifies header alignment or enforcing plausible value ranges would have rejected the malformed parse at the interface boundary.
Instead, structurally corrupted but locally plausible data propagated across modules,
increasing execution cost and obscuring the root cause.
\end{example}

\begin{figure*}[t]
    \centering
    \includegraphics[width=0.72\linewidth]{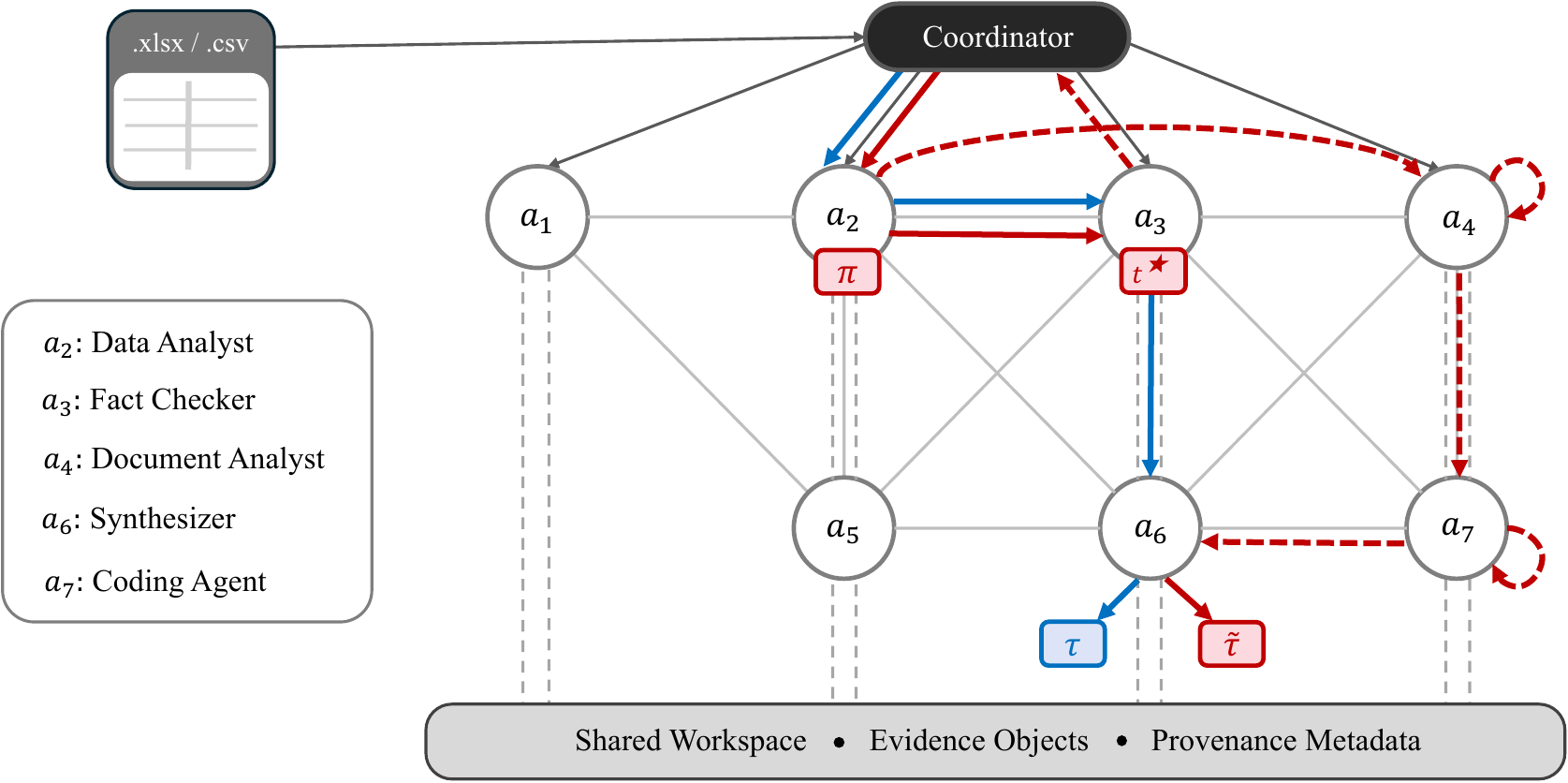}
    \caption{An illustrative failure mode in a multi-agent workflow analyzing quarterly revenue data. A table parsing error 
    introduced by perturbation operator $\pi$ causes downstream computations to produce nonsensical growth rates. Rather than 
    failing immediately, the workflow reroutes (divergence point $t^\star$) to propose alternative interpretations, retries 
    extraction with different settings, and explores multiple table structure hypotheses, expanding execution from 3 to 9 steps. 
    We record execution traces from both the clean ($\tau$) and perturbed ($\tilde{\tau}$) runs to track the divergence of the 
    execution trajectory.}
    \label{fig:workflow_structure}
\end{figure*}

We study uncertainty propagation by treating uncertainty as a controlled experimental variable. We inject structured perturbations into 
artifact-derived representations (e.g., extracted text spans and tables) under varying perturbation types, execute fixed workflows under 
comprehensive execution logging, and quantify contamination using trace divergence
i.e., the extent to which plans, tool invocations, evidence selection, 
intermediate state, and inter-agent messages deviate from a noise-free baseline. This trace-level framing makes propagation measurable: it 
identifies where divergence begins, how far it spreads, and which interfaces and decision boundaries are most sensitive to upstream uncertainty.

Our experiments instantiate structured workflows as a multi-agent orchestration and evaluate on \numTasks GAIA 
tasks with file attachments across tabular, document, image, and audio modalities, analyzing \numRuns paired 
clean/perturbed runs across \numLLMs language models (GPT-5-mini, LLaMA-3.1-70B, Qwen3-235B). Through trace-level analysis, we uncover that \emph{structural divergence 
and outcome corruption are decoupled}. Workflows may diverge substantially yet recover correct answers 
(behavioral detours with recovery, \pctBehavioralDetours of runs), or remain structurally similar while producing incorrect 
outputs (silent semantic corruption, \pctSilentCorruption of runs). This challenges outcome-only evaluation: answer accuracy 
misses costly internal instabilities, while trace divergence can reflect healthy adaptation.

We identify recurring contamination patterns—strategy rerouting (\pctRerouting of divergent runs), extended execution 
(\pctExtendedExecution), and early termination (\pctEarlyTermination) and show they exhibit characteristic temporal signatures (first 
divergence point) and modality-specific fingerprints. Tabular perturbations predominantly trigger extended 
execution (24.4\%); audio perturbations favor early termination. Behavioral detours consume median 
\costBehavioralDetours baseline tokens (IQR: 1.1--2.5$\times$), while silent semantic corruption exhibits near-baseline cost, 
revealing a cost-correctness tradeoff. 
In practice, this means many high-cost runs are not failures, 
and many low-cost runs are not trustworthy.

We contribute: (i)~a formal taxonomy of contamination manifestations (silent semantic corruption, behavioral 
detours, combined disruption) and their control-flow signatures in structured workflows; (ii)~a trace-based 
measurement framework for detecting and localizing contamination via structural divergence, first divergence 
point, and operational cost metrics; and (iii)~empirical evidence and design insights from a 
multi-agent orchestration evaluated on \numRuns runs across GAIA tasks, with implications for cost-aware verification, 
targeted hardening, and why common guardrails fail to intercept contamination. Artifacts available at \href{https://github.com/anna-mazhar/trace-level-contamination-mas}{the repository}.
\vspace{-0.04in}
\section{Background and Related Work}
\label{sec:related}

We review tool-augmented architectures, coordination mechanisms, 
and evaluation methods, then examine existing approaches to uncertainty, 
debugging, and verification—revealing a key gap: 
how information corruption propagates through agent workflows and 
evades current safeguards.

\vspace{-3pt}

\paragraph{Tool-augmented agent architectures.}
Language model agents increasingly incorporate external tools to overcome 
limitations in knowledge, computation, and grounding. 
Early works like Toolformer \cite{toolformer2023} demonstrated that LLMs can learn 
when and how to invoke APIs for calculator operations, retrieval, 
and translation, while ReAct \cite{yao2023react} introduced 
interleaved reasoning traces and tool actions, enabling more 
interpretable and grounded multi-step task solving. Subsequent 
systems such as PAL \cite{pmlr-v202-gao23f}, 
ART \cite{paranjape2023artautomaticmultistepreasoning}, and 
TALM \cite{parisi2022talmtoolaugmentedlanguage}
extended this pattern through code execution, 
explicit decomposition, and improved tool-use reliability. 
While effective on complex tasks, 
these architectures introduce sequential dependencies in 
which early errors can compound downstream.

\vspace{-3pt}
\paragraph{Multi-agent systems and coordination.}
Multi-agent systems distribute tasks across specialized modules 
when a single agent is insufficient. Representative frameworks 
include AutoGen \cite{wu2024autogen}, MetaGPT \cite{hong2024metagpt}, 
ChatDev \cite{qian-etal-2024-chatdev}, and CAMEL \cite{NEURIPS2023_a3621ee9}, 
while more dynamic orchestration strategies appear in Magentic-One 
\cite{fourney2024magenticonegeneralistmultiagentsolving}, 
Mixture-of-Agents \cite{wang2024mixtureofagentsenhanceslargelanguage}, 
and Captain Agent \cite{song2025adaptiveinconversationteambuilding}. 
These systems demonstrate strong capabilities, 
but their evaluation primarily emphasizes end-task success rather than 
information propagation across agents.
\vspace{-3pt}
\paragraph{Uncertainty and robustness in agent workflows.}
Robustness testing for LLM-based systems has focused primarily on input perturbations and adversarial attacks.
CheckList \cite{ribeiro-etal-2020-beyond} introduced behavioral testing for NLP models, systematically probing capabilities and failure modes.
Recent work examines prompt robustness: PromptRobust \cite{10.1145/3689217.3690621} evaluates LLMs under adversarial prompt perturbations, 
while Jailbreak attacks \cite{NEURIPS2023_fd661313} explore safety vulnerabilities through carefully crafted inputs.
In the context of retrieval-augmented generation, work on RAG robustness \cite{Chen_Lin_Han_Sun_2024} studies how noise in retrieved 
documents affects generation quality.
However, these efforts concentrate on single-model robustness or end-to-end task performance.
In broader machine learning pipelines, error propagation has been studied 
in the context of uncertainty quantification \cite{ABDAR2021243}, 
where distributional assumptions allow tracking confidence degradation across model cascades.
In software systems, cascading failures have been extensively analyzed in 
distributed systems and microservices \cite{10.5555/1251460.1251461}.
Our work bridges these perspectives, treating multi-agent workflows as systems 
where information flows across loosely-coupled modules. 
\vspace{-3pt}
\paragraph{Evaluation and benchmarking of agent systems.}
Agent benchmarks assess performance on diverse reasoning tasks.
GAIA \cite{mialon2024gaia} provides real-world tasks that require 
multi-step reasoning over diverse file types, including PDFs, spreadsheets, and images. 
Agent benchmarks more broadly, such as AgentBench \cite{liu2025agentbenchevaluatingllmsagents} (multi-env agent tasks), 
WebArena \cite{zhou2024webarenarealisticwebenvironment} and Mind2Web \cite{NEURIPS2023_5950bf29} (web navigation and interaction), 
and SWE-bench \cite{jimenez2024swebenchlanguagemodelsresolve} (software issue resolution), 
assess performance across a range of environments and task settings, 
typically reporting success/failure and cost. 
Other efforts, including AgentBoard \cite{agentboard2024}, 
Agent Lumos \cite{agentlumos2024}, and 
MAST \cite{cemri2025multiagentllmsystemsfail} 
(subtask progress tracking, reasoning-chain supervision, and failure taxonomy, respectively),
move toward finer-grained progress tracking and failure taxonomy, 
but still do not directly characterize contamination propagation through execution traces.

\vspace{-3pt}
\paragraph{Debugging and introspection in agent systems.}
As agent systems grow in complexity, debugging and introspection 
have become increasingly important.
Observability and optimization tools such as 
LangSmith \cite{langchain_langsmith} and DSPy \cite{khattab2023dspy} 
support tracing and systematic pipeline refinement, 
while related work has explored explainability \cite{zhao2024explainability}, 
self-debugging \cite{reflexion2023}, hierarchical debugging \cite{autodan2023}, 
and automated failure attribution \cite{zhang2025agentcausestaskfailures}.
However, these approaches are largely post hoc: they help diagnose 
failures after they occur, rather than systematically 
identifying which perturbations lead to which contamination behaviors.

\vspace{-3pt}
\paragraph{Verification and guardrails.}
Ensuring safe and reliable agent behavior has motivated a range of verification 
and guardrail approaches, including principle-based self-critique \cite{constitutionalai2022}, 
programmatic input/output validation \cite{guardrailsai} including format validation, 
semantic checks, and toxicity filters.
Other methods focus on LLM uncertainty estimation to trigger fallback behaviors, 
tool-call checking, and formal verification for generated programs 
\cite{semanticuncertainty2023, austin2021programsynthesislargelanguage}.
While valuable, these methods typically operate on individual outputs or final outcomes 
and therefore provide limited visibility into how locally plausible 
but corrupted information propagates across agent interactions.
A sanitized but incorrect extraction from Agent A 
may pass local checks yet still contaminate Agent B's reasoning.

\vspace{-3pt}
\paragraph{Positioning our work.}
While prior work has established powerful agent architectures, evaluated their task-level performance, 
and developed verification mechanisms, 
a key gap remains: \emph{understanding how uncertainty propagates through agent workflows}.
We contribute a trace-based methodology for controlled experimentation, a taxonomy of contamination mechanisms 
observed in multi-agent systems, and empirical evidence that 
existing guardrails often fail to catch these failures.
Our work provides a foundation for designing propagation-aware verification strategies 
and more robust agent coordination protocols.

\vspace{-6pt}
\section{Problem setup and definitions}
\label{sec:definitions}

We study structured \emph{multi-agent workflows} in which a task is decomposed across specialized
agents that exchange messages and invoke tools over heterogeneous artifacts (PDFs, spreadsheets,
slide decks, etc.). Figure~\ref{fig:workflow_structure} shows the workflow structure
of the revenue analysis example from \S\ref{para:example_failure_mode} which we will use to
ground the formal definitions below.

The workflow is represented as a directed interaction graph
$\mathcal{G}=(\mathcal{A},\mathcal{E})$, where each node $a\in\mathcal{A}$ denotes an agent with a
role-specific policy and toolset. Each directed edge $(a_i,a_j)\in\mathcal{E}$ indicates that
agent $a_j$ may consume messages or tool outputs produced by $a_i$. Within the graph, agent $a_i$ is
\emph{upstream} to $a_j$ if there is a directed path from $a_i$ to $a_j$; correspondingly, $a_j$ is
\emph{downstream} from $a_i$. Information is \emph{upstream} when produced by agents earlier in the
execution DAG, and information is \emph{downstream} when consumed by agents later in the DAG.

We ground the formal definitions below using the quarterly revenue analysis scenario from \S\ref{para:example_failure_mode}:
a corrupted table parse forces downstream routing decisions and expands execution from 3 to 9 steps.
Figure~\ref{fig:workflow_structure} depicts both the clean and perturbed execution traces, marking the 
first divergence point $t^\star$ and propagation pattern.

\vspace{-6pt}
\paragraph{Execution traces.}
A single workflow run induces an \emph{execution trace} $\tau=(e_1,\dots,e_T)$, an ordered sequence
of logged workflow events. In our implementation, events are drawn from a fixed schema including
routing decisions (which agent is selected next), tool invocations (parse table, validate schema, etc.), 
memory reads/writes, retrieval displays, agent outputs, and the task outcome event. Each event carries 
a typed payload (e.g., selected agent, tool name and operation, success/failure flag, memory entry type, 
action type). Figure~\ref{fig:workflow_structure} depicts both the clean and perturbed execution traces.

\vspace{-6pt}
\paragraph{Structural event signatures.}
To compare traces robustly despite lexical variation (different wordings, timestamps, content hashes),
we abstract each event to a \emph{structural signature} which is a compact representation preserving only
control-flow-relevant information. For example, when the analyst agent in the revenue scenario makes a 
routing decision, the signature records \emph{which} agent it selected 
(e.g., proceed with analysis or reroute to validation), not the LLM reasoning that led to the decision. 
Similarly, the signature of the table parsing tool invocation records the tool name, 
operation, and success/failure status, but not the full parsed output. 
Formally, we map each event $e_t$ to a \emph{structural signature}
$\sigma(e_t)$. The signature sequence for a trace is
\[
S(\tau) = (\sigma(e_1),\dots,\sigma(e_T)).
\]

\paragraph{Perturbations.}
A \emph{perturbation} is a controlled transformation applied to an upstream information item
$x$ before downstream consumption. Let $\pi$ denote a perturbation operator and
$\tilde{x}=\pi(x)$ its perturbed version. A perturbed run produces a trace $\tilde{\tau}$ in which
one or more consumed items are replaced by perturbed counterparts. Examples include table column swaps, 
OCR noise in documents, and image blurring (detailed in Methodology). 
These reflect realistic failure modes, as in the revenue scenario, 
the table parser misidentifies merged header cells, 
causing downstream queries to reference wrong columns.

\vspace{-6pt}
\paragraph{Trace divergence.}
We quantify divergence by comparing the structural signature sequences $S(\tau)$ and
$S(\tilde{\tau})$ using edit distance under minimum-edit alignment 
(Wagner--Fischer dynamic programming), which yields substitutions, insertions, 
and deletions between the two signature sequences. 
Our primary trace-level divergence metric is the normalized structural edit distance
\[
d_{\mathrm{norm}}(\tau,\tilde{\tau})
=
\frac{\mathrm{ED}(S(\tau),S(\tilde{\tau}))}{\max(|S(\tau)|,|S(\tilde{\tau})|)}.
\]
This metric ranges from 0 (identical execution patterns) to 1 (completely different traces).
In the revenue scenario, the clean trace signature sequence is compact (roughly 3--4 events), 
while the perturbed trace stretches to 9 with inserted validation and rerouting operations, 
yielding a substantial normalized divergence.

\vspace{-6pt}
\paragraph{First divergence point and cascade summaries.}
The overall edit distance measures total disruption, but for diagnosis we need to pinpoint \emph{where}
divergence begins. Under the edit-distance-induced alignment, the \emph{first divergence point} $t^\star$ 
is the earliest aligned event index at which the structural signatures differ. We record not just the 
timing, but also the type of first divergence (e.g., reroute, tool mismatch, action mismatch). This 
locates the boundary crossing and decision point most immediately affected by upstream corruption.
Figure~\ref{fig:workflow_structure} marks $t^\star$ for the revenue scenario, 
where the first divergence is a routing decision to reroute to validation 
rather than proceed with analysis.

\smallskip\noindent\textbf{Problem statement.}
Multi-agent workflows increasingly rely on externally derived information (extracted text from 
documents, parsed tables, computed values from tools) to make downstream routing and reasoning 
decisions. Errors or uncertainties introduced during information extraction and tool execution 
can propagate through agent boundaries, compounding across steps and leading to incorrect task outcomes 
or inefficient execution paths. We study how controlled perturbations to upstream 
information affect multi-agent workflow behavior, aiming to characterize the patterns and 
severity of information contamination cascades, hoping to inform the design of more robust 
and interpretable multi-agent systems.

\vspace{-0.04in}
\section{Experimental Setup}
\label{sec:experiment_setup}
This section describes our trace-centric measurement approach 
and its instantiation on the GAIA benchmark. 
We execute paired clean and perturbed workflows using 
formally defined execution traces (\S\ref{sec:definitions}), 
log intermediate artifacts with provenance, and quantify contamination 
via structural divergence and outcome measures.

\smallskip\noindent\textbf{Benchmark and Task Selection.}
\label{subsec:task_selection}
We evaluate on the GAIA benchmark, selecting \numTasks tasks that include one or more file attachments. 
Common attachment types include PDFs, DOCX, PPTX, XLSX/CSV tables, images, 
and audio files. Each task retains its original prompt and attachment bundle. This diversity of modalities 
and tasks enables us to observe contamination patterns across heterogeneous reasoning primitives.

\vspace{-0.05in}

\subsection{Multi-Agent Orchestration}
\label{subsec:agent_instantiation}

We instantiate the workflow as a coordinated multi-agent system with the following design:

\smallskip\noindent\textbf{Architecture.} 
A small set of specialized agents (extraction, analysis, code generation, validation, etc.) 
communicate through a shared workspace. A coordinator (LLM-based router) selects which agent 
to invoke next based on the current task state and workspace contents. This apparatus is 
\emph{experimental} (not a proposed contribution) and is held fixed across clean and perturbed 
conditions to enable fair paired comparisons. The apparatus design ensures multiple 
information-exchange boundaries and heterogeneous reasoning primitives (parsing, tabular 
manipulation, computation, synthesis), allowing us to observe where perturbed evidence 
crosses boundaries and how downstream decisions respond.

Details on agent roles, memory schema, and orchestration architecture are 
in Appendix~\ref{subsec:appendix_architecture}. 

\smallskip\noindent\textbf{Instrumentation and Provenance Tracking.}
\label{subsec:instrumentation}
For each run, we record a structured event trace (\S\ref{sec:definitions}) capturing all routing 
decisions, tool invocations, memory operations, agent outputs, and the task outcome.
Beyond the event trace, we track artifact provenance. Each logged output (tool result, memory entry, 
or agent message) records its upstream dependencies. This dependency graph enables us to identify 
which downstream artifacts depend on perturbed information, critical for contamination scoping.

\smallskip\noindent\textbf{Modality-Aware Perturbation Operators.}
\label{subsec:perturbation_operators}
Following the formal perturbation model from \S\ref{sec:definitions}, we apply perturbation 
operators $\pi$ to artifact-derived representations (e.g., extracted tables, parsed text), 
rather than modifying raw files. This reflects realistic failure modes (extraction and parsing errors) 
at the points where agents consume information.
In brief, we perturb tabular, document, image, and audio attachments with modality-matched operators that induce content, and structural corruption.
More details and rationale for each operator are in Appendix~\ref{subsec:appendix_perturbation} and~\ref{subsec:appendix_perturbation_types_rationale}, where we also summarize the modality-specific operator set.

\vspace{-0.03in}

\subsection{Controlled Variables and Reproducibility}
\label{subsec:reproducibility}

We hold constant all non-perturbed variables across paired runs:

\smallskip\noindent\textbf{Execution control.}
Agent roles, prompts, tool wrappers, shared-state schema, stopping/retry policies, and random seeds are fixed.

\smallskip\noindent\textbf{Logging and seeding.}
For each run, we record perturbation type, injection locus, and affected evidence identifiers
using fixed random seeds. 
On a subset of 20 tasks, repeating clean runs five times yielded low baseline 
trace variation overall (pairwise normalized structural edit distance: 
median 0.0, IQR 0.1173; mean 0.0501). 
All runs use a single LLM backend at temperature $0$ to minimize 
sampling variance. Exact tool wrappers 
and library versions are documented in the appendix.

\vspace{-0.03in}

\begin{table}[t]
\centering
\small
\caption{Metrics used in the main analysis. We prioritize interpretable trace-level measures 
and treat detailed event payloads as implementation-level diagnostics.}
\setlength{\extrarowheight}{2pt}
\begin{tabular}{>{\raggedright\arraybackslash}p{0.35\linewidth} 
                >{\raggedright\arraybackslash}p{0.6\linewidth}}
\toprule
\textbf{Metric} & \textbf{Role in analysis} \\
\midrule
\textbf{Structural edit distance} 
& Trace-level disruption score; comparable across tasks and perturbations \\[15pt]

\textbf{First divergence point} 
& Identifies when execution first deviates ($t^\star$ timing) \\[15pt]

\textbf{Control-flow pattern prevalence} 
& Quantifies rerouting, looping/extended execution, and early termination \\[15pt]

\textbf{Control-flow diagnostics} 
& Captures tool-call changes, retries, failures, and truncation/extension behavior \\[15pt]

\textbf{Task success} 
& End-task robustness under perturbation \\[5pt]

\textbf{Token overhead} 
& Relative cost (perturbed vs. clean) under retries, detours, and failure loops \\
\bottomrule
\end{tabular}
\label{tab:main-metrics}
\end{table}

\vspace{-0.03in}

\subsection{Trace Divergence and Outcome Metrics}
\label{subsec:evaluation_metrics}

We quantify contamination using metrics defined formally in \S\ref{sec:definitions}
and summarized in Table~\ref{tab:main-metrics}.

\subsubsection{Trace-level metrics}
\label{subsec:trace_metrics}

We report three trace-level metrics to characterize structural divergence, defined in \S\ref{sec:definitions}:

\begin{itemize}
\item \textbf{Structural edit distance} $d_{\mathrm{norm}}(\tau,\tilde{\tau})$: the primary divergence score
\item \textbf{First divergence point} $t^\star$: the timing of initial deviation
\item \textbf{Control-flow diagnostics}: including reroutes, added/removed tool calls, introduced failures, early termination, and extended execution
\end{itemize}

\subsubsection{Outcome metrics}
\label{subsec:outcome_efficiency}

We evaluate task outcome using benchmark-appropriate scoring. 
We record whether the task outcome changed and measure execution cost primarily via token overhead
(perturbed vs. clean), with step/tool-invocation changes treated as supporting diagnostics.
These capture expensive cascades (retries, detours, loops) that divergence alone may not reflect.

\vspace{-0.04in}
\section{Manifestation Patterns}
\label{sec:manifestations}
We analyze \numRuns paired clean/perturbed runs across \numTasks GAIA validation set tasks, applying modality-specific
perturbation operators to artifact-derived representations (\S\ref{sec:experiment_setup}).  
Our goal in this section is to characterize \emph{how}
uncertainty injected into artifact-derived information manifests in structured workflows---as
execution-level disruption, outcome corruption, or both---and to extract recurring mechanisms
that support debugging and targeted mitigation. While we collected data across three LLM backends 
(GPT-5-mini, LLaMA-3.1-70B, Qwen3-235B), the analysis below focuses on GPT-5-mini; results for 
LLaMA and Qwen are provided in Appendix~\ref{sec:appendix_llm_results}.

\subsection{Divergence vs.\ Outcome Corruption}
\label{subsec:divergence_vs_outcome}

A central observation from our trace-level analysis is that contaminated information does not 
always manifest as task failure. Perturbations trigger two related but distinct forms of disruption: 
\emph{structural divergence} (changes in agent sequencing, tool calls, and execution paths) and 
\emph{outcome corruption} (changes in task outcomes). Critically, these dimensions are decoupled: 
workflows may diverge substantially yet recover correct answers, or remain structurally similar 
while producing incorrect outputs.

We define \emph{recovery} as a perturbed run producing the same task outcome as the clean baseline, 
even if its intermediate execution diverges structurally. This concept is critical for understanding 
that workflows may exhibit internal instability while still producing correct results.

\begin{figure}[t]
\centering
\includegraphics[width=1.0\columnwidth]{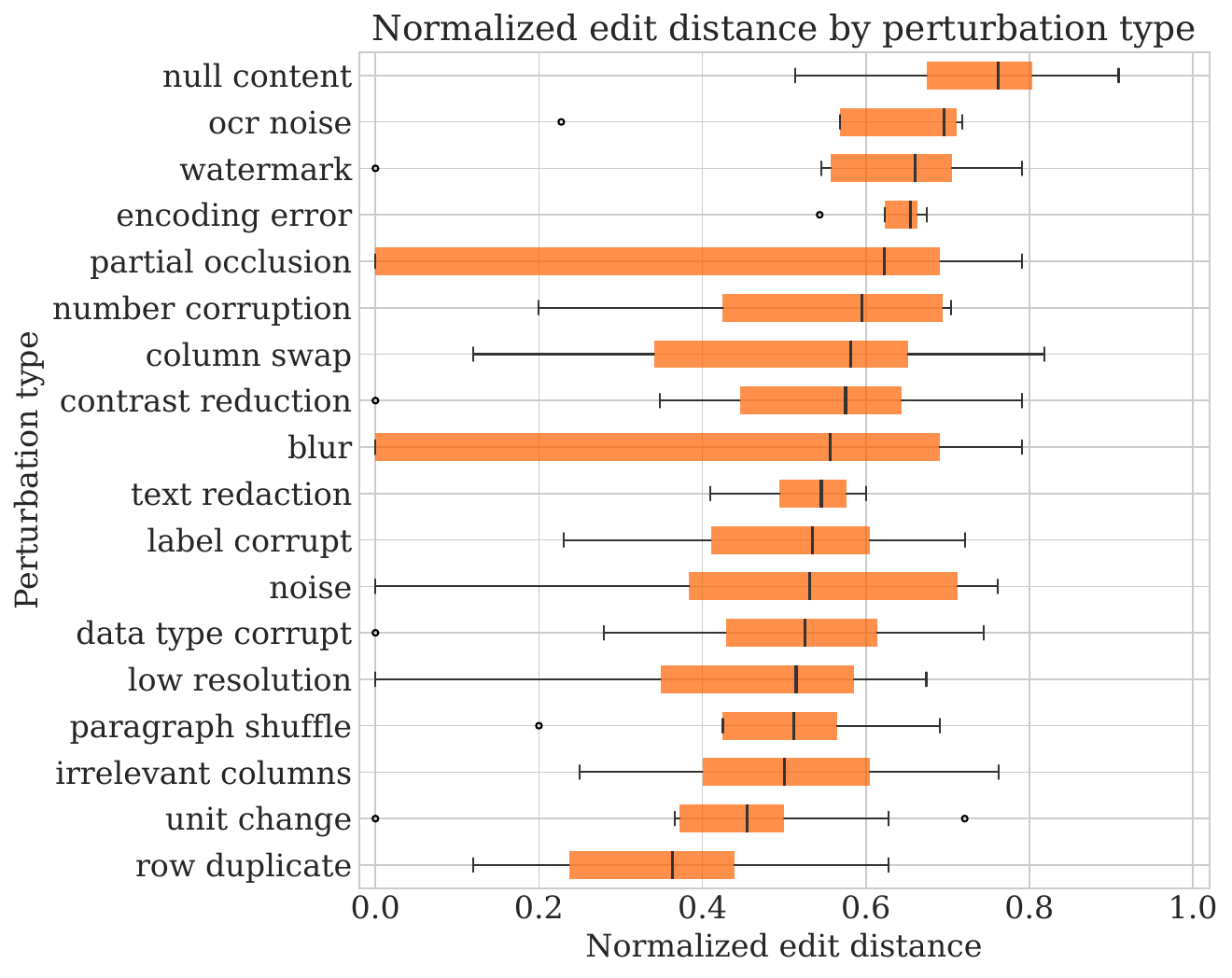}
\caption{Structural edit distance by perturbation type. OCR noise induces 
consistent structural change with low variation, while image blur exhibits high variance, 
revealing differential adaptive responses.}
\label{fig:edit_distance}
\end{figure}
\begin{figure}[t]
\centering
\includegraphics[width=1.0\columnwidth]{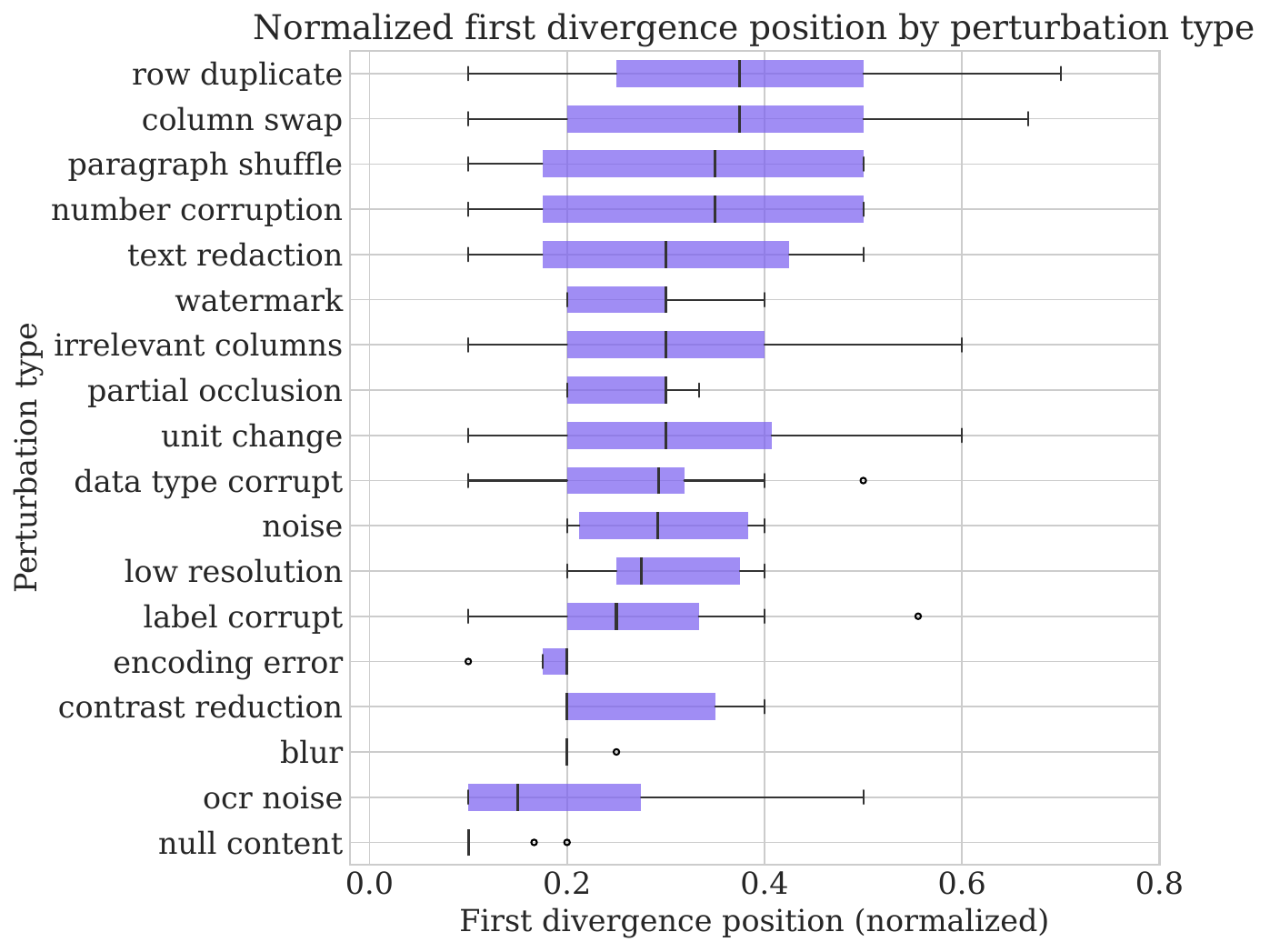}
\caption{First divergence point by perturbation type.
Section removal perturbations in documents show the most frequent earliest divergence, 
while OCR noise exhibits least variance, suggesting different intensities of contamination.}
\label{fig:divergence_timing}
\end{figure}
\begin{figure}[t]
\centering
\includegraphics[width=1.0\columnwidth]{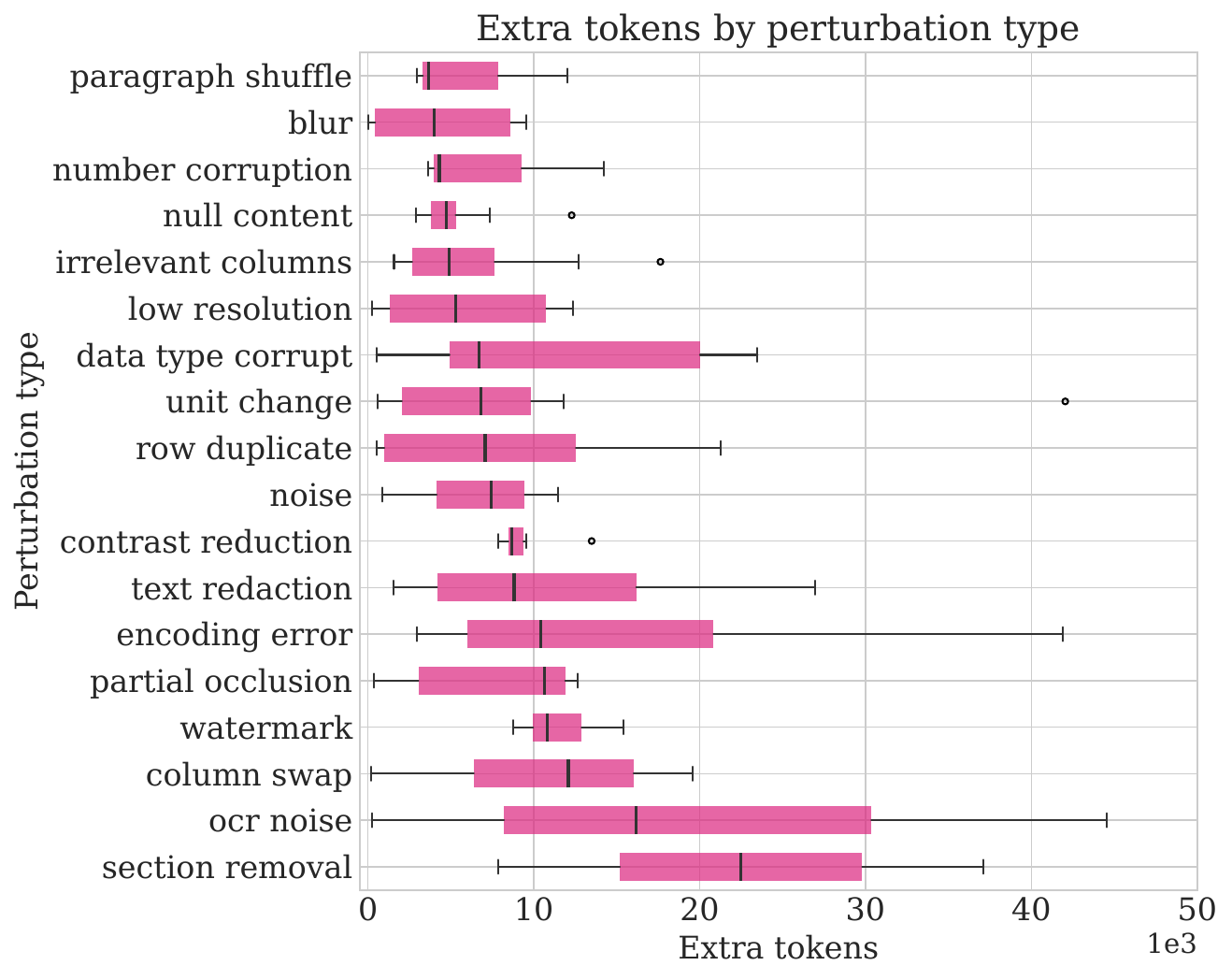}
\caption{Token overhead by perturbation type. Contrast reduction in images 
and number corruption show consistent overhead patterns, 
while data-type corruption in tabular data exhibits higher variance.}
\label{fig:token_overhead}
\end{figure}

\textbf{Silent semantic corruption.}
Structurally, the perturbed trace $\tilde{\tau}$ exhibits a signature sequence $S(\tilde{\tau})$ 
nearly identical to the clean baseline $S(\tau)$: the routing events, tool invocation events, and 
agent output events align closely, yielding structural edit distance $d_{\mathrm{norm}}(\tau,\tilde{\tau}) \approx 0$, 
 as shown in Figure~\ref{fig:edit_distance}. 
However, despite this close alignment in control-flow events, 
the task outcome event $e_T$ can still differ between runs. This pattern arises when perturbations 
introduce subtle semantic shifts (e.g., off-by-one cell references, unit mismatches, or corrupted 
numeric values) that propagate through the workflow without triggering changes in the control-flow 
events $e_t$.

This was frequently observed in tasks with image attachment. When watermark perturbation was applied, both clean and 
perturbed executions produced identical routing events (planner $\to$ visual analyst $\to$ 
synthesizer) and identical tool invocation signatures, yet the task outcome event $e_T$ recorded 
different outputs (clean run produced a longer list of fractions; perturbed run omitted entries). 
The watermark shifted the image representation, altering visual interpretation without affecting 
the structural event sequence. This illustrates the core challenge: contaminated information is 
consumed (visible in provenance dependencies) and manifests in $e_T$, but the trace signature 
sequence remains nearly identical. From a debugging perspective, these failures are particularly 
insidious: outcome-level validation detects the error, but structure-level trace comparison 
provides limited localization signal upon first divergence point analysis.

\textbf{Behavioral detours with recovery.}
The perturbed trace $\tilde{\tau}$ diverges substantially from the clean trace $\tau$, exhibiting different 
routing events, additional tool invocation events, or reordered agent output events, yet the task 
outcome event $e_T$ matches the clean baseline. This pattern reflects adaptive behavior: the 
workflow encounters corrupted information, takes an alternative execution path through modified 
$S(\tilde{\tau})$, and successfully compensates through redundancy, cross-checking, or fallback 
strategies.

For instance, in a task with spreadsheet attachment, data-type corruption (symbols injected into 
numeric cells) was applied. The perturbation induced substantial structural divergence: execution 
expanded from 3 to 9 steps, introducing additional routing cycles and tool invocations (repeated 
Python executions, fact-checking passes). The first divergence point $t^\star$ occurred early, as 
corrupted numeric fields disrupted standard parsing and aggregation. Nevertheless, both runs 
produced identical task outcomes, demonstrating successful recovery despite noisy inputs. While the 
outcome is correct, the divergence reveals brittleness in the nominal execution path and carries 
significant cost implications: the perturbed run consumed substantially more steps and tool 
interactions (see \S{\ref{subsec:operational_cost}}). With input-sanitization mechanisms, much of this additional execution cost could be 
avoided.

This decoupling has methodological consequences. Outcome-level robustness metrics (answer accuracy 
under perturbation) can miss meaningful contamination when internal behavior is unstable but the 
system recovers through alternate paths. Conversely, trace-based divergence metrics can overstate 
harm when the workflow adapts successfully. A complete characterization therefore requires both 
behavioral and outcome views. For multi-agent workflows, where multiple valid execution paths may 
solve the same task, this dual perspective is essential: structural divergence may reflect healthy 
adaptation rather than failure, and semantically consequential errors may occur through localized 
changes without broad control-flow disruption.

\begin{figure}[t]
\centering
\includegraphics[width=0.84\columnwidth]{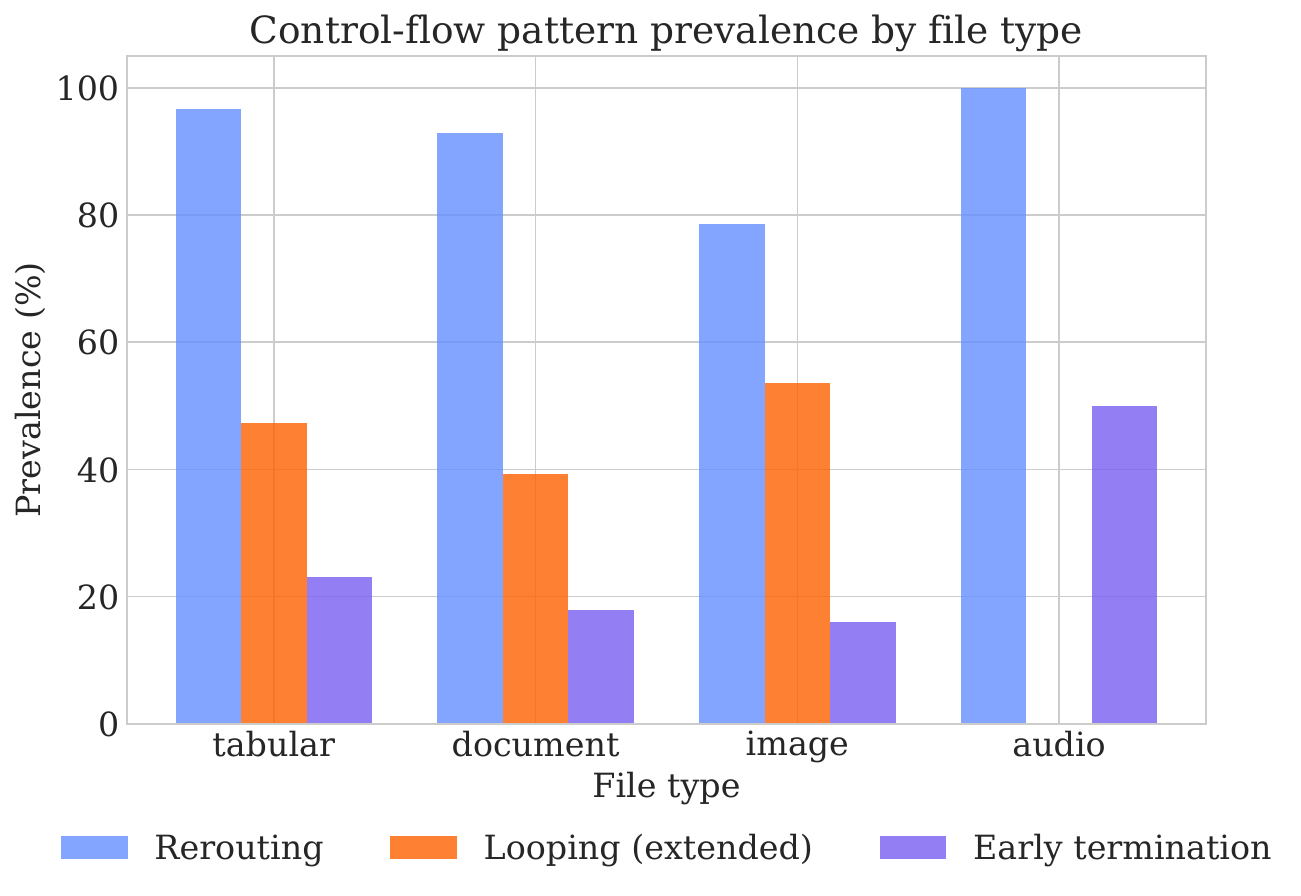}
\caption{Control-flow patterns (rerouting, looping, termination) by artifact modality.}
\label{fig:control_flow_by_file_type}
\end{figure}

\begin{figure}[t]
\centering
\includegraphics[width=1.0\columnwidth]{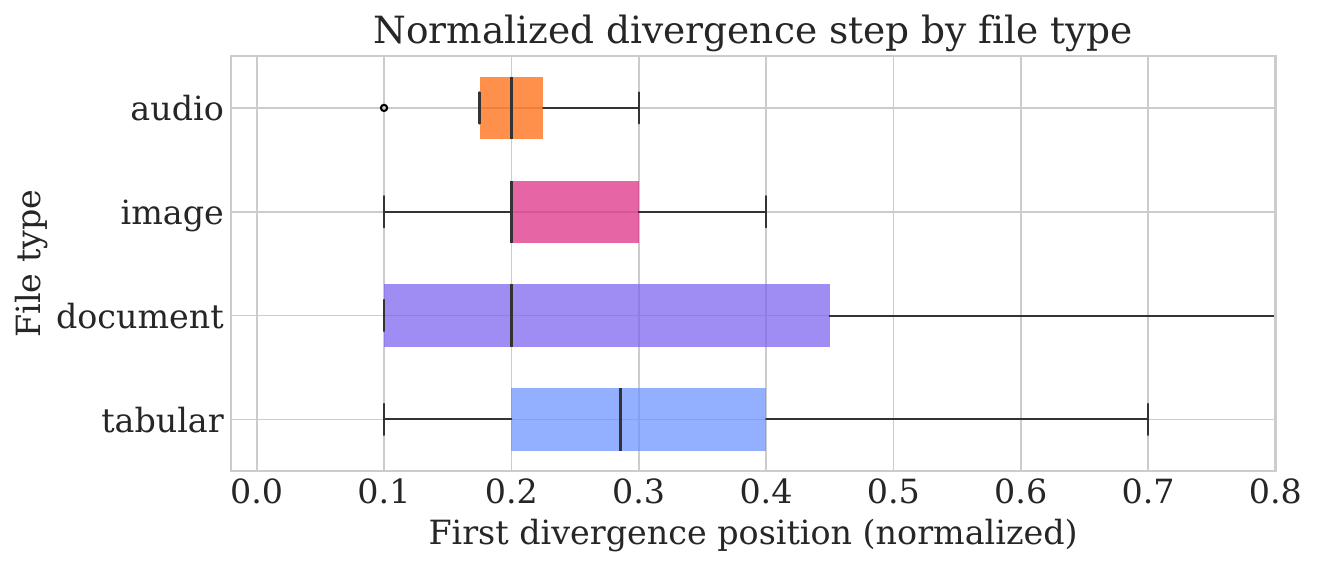}
\caption{First divergence point timing by artifact modality.}
\label{fig:divergence_step_by_file_type}
\end{figure}

\textbf{Structural disruption with outcome corruption.}
In this case, both structure and outcome differ, representing the most severe contamination regime. 
Here, contamination cascades beyond what adaptive strategies can mitigate; 
recovery fails because the system lacks either adequate tools or 
sufficient model reasoning capability.

\smallskip\noindent\textbf{Prevalence.}
Across all perturbed runs, \pctSilentCorruption exhibit silent semantic corruption, \pctBehavioralDetours show 
behavioral detours with recovery, and 39.9\% exhibit both structural and outcome corruption 
(combined disruption). The rest of the runs (4.5\%) show neither structural nor outcome disruption, 
indicating perturbations that were effectively ignored or had no impact.

\finding{Outcome-only metrics miss substantial internal contamination. Workflows frequently recover 
correct answers despite major structural divergence (40.3\%), or silently fail while maintaining stable 
traces (15.3\%). This makes endpoint-only metrics inadequate.}

\vspace{-0.03in}
\subsection{Structural Control-Flow Patterns}
\label{subsec:structural_patterns}

When perturbations induce structural divergence, they follow recurring control-flow patterns that reveal 
distinct failure modes and localize vulnerabilities to specific workflow components:

\paragraph{Strategy rerouting.}
Different agents are selected, alternative tools are invoked, or reasoning steps are reordered. 
This pattern suggests contamination affects routing decisions or confidence calibration. 
\pctRerouting of divergent runs exhibit rerouting as the primary signature.

\paragraph{Extended execution and looping.}
The perturbed run requires additional routing cycles, retries, or detours. This arises when tool 
outputs become ambiguous or inconsistent, triggering retry logic or multi-stage verification. 
\pctExtendedExecution of divergent runs exhibit extended execution. From a cost perspective, these runs 
consume disproportionate resources (median overhead: 2.4$\times$ baseline tokens).

\paragraph{Early termination.}
The perturbed run halts prematurely, skipping downstream agents or synthesis steps. This emerges 
when perturbations cause parsing failures, empty tool outputs, or confidence thresholds triggering 
early exit. \pctEarlyTermination of divergent runs terminate early, often leading to incomplete answers.

Figure~\ref{fig:control_flow_by_file_type} summarizes pattern prevalence by file type.
A single run may exhibit multiple patterns sequentially (e.g., rerouting followed by looping).
Rerouting dominates across all modalities: when agents detect inconsistencies, 
they trigger alternative analysis strategies.
Audio exhibits a distinctive early termination pattern, 
where the audio agent halts execution when transcription fails, rendering further processing infeasible.

\finding{Contamination exhibits modality-specific failure signatures. Rerouting dominates across 
tabular and document perturbations (\pctRerouting of divergent runs). Audio uniquely favors early termination, 
where failed transcription halts downstream processing entirely. These fingerprints enable targeted defense.}

These structural patterns provide actionable localization signals. Rerouting-heavy perturbations 
localize failures to routing policy decisions and confidence calibration components; loop-heavy 
perturbations localize issues to retry logic and stopping criteria; termination-heavy perturbations 
localize gaps to failure recovery and fallback mechanisms. Robustness claims based solely on 
terminal accuracy understate the prevalence of contamination by missing these internal disruptions.

\vspace{-0.03in}
\subsection{Temporal Localization}
\label{subsec:divergence_timing}

The first divergence point $t^\star$ (normalized position in trace) reveals when contamination 
manifests. First divergence point is not uniform: some perturbations trigger immediate divergence; 
others manifest after several apparently normal steps.
Figure~\ref{fig:divergence_timing} and Figure~\ref{fig:divergence_step_by_file_type} show 
first divergence point distributions across different modalities and perturbation types.

\paragraph{Early divergence.}
Perturbations that cause early divergence (median $t^\star / T < 0.1$) 
typically disrupt initial interpretation, parsing, or 
grounding. For example, severe structural corruptions (e.g., column misalignment, encoding errors) 
may prevent agents from reliably reading input artifacts, triggering immediate rerouting or failure. 
Early divergence signals that the workflow cannot establish a stable foundation for downstream 
reasoning.

\paragraph{Late divergence.}
Perturbations that cause late divergence (median $t^\star / T > 0.3$) indicate that early processing remains intact but 
contamination is exposed when the workflow reaches subsequent reasoning, computation, or synthesis 
stages. For instance, a subtle numeric corruption pass initial extraction and validation but 
cause divergence when an agent performs arithmetic comparison or constraint checking. Late 
divergence is informationally valuable: it localizes which part of the pipeline is most sensitive 
to the perturbation and must be targeted for verification. 

We found that first divergence timing also varies by modality 
(Figure~\ref{fig:divergence_step_by_file_type}): audio perturbations trigger early, consistent divergence, 
while document perturbations exhibit high variance—reflecting broader attack surface and 
diverse processing stages.

\finding{First divergence timing reveals failure mechanism. Early divergence ($t^\star < 0.1T$) signals 
foundational extraction failures; late divergence ($t^\star > 0.3T$) reveals reasoning-stage sensitivity. 
This temporal signature guides where to harden verification: early for parsing, late for computation.}

These patterns demonstrate that a simple ``severity'' framing is insufficient. 
Perturbations do not vary only in \emph{how much} they disrupt execution; they differ qualitatively 
in \emph{mechanism}: whether they primarily alter routing decisions, induce retries, change tool 
success states, truncate workflows, or silently shift semantic content. Understanding \emph{how} a 
perturbation disrupts execution enables targeted fixes (e.g., improving specific tool robustness, 
adjusting confidence thresholds, or hardening particular agent transitions), and explains apparent 
inconsistencies in aggregate outcome metrics where perturbations with similar answer-change rates 
may produce vastly different operational costs and trace patterns.

\vspace{-0.06in}
\begin{figure}[t]
\centering
\includegraphics[width=1.0\columnwidth]{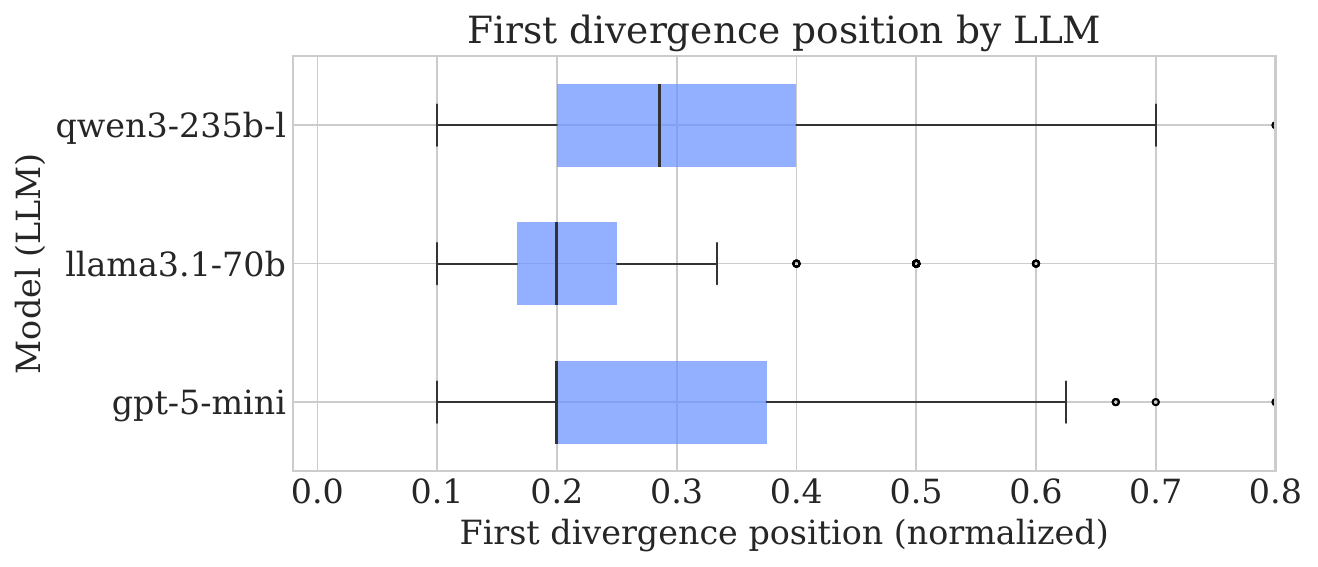}
\caption{First divergence point timing by LLM backend.}
\label{fig:divergence_step_by_llm}
\end{figure}

\vspace{-0.06in}
\begin{figure}[t]
\centering
\includegraphics[width=0.8\columnwidth]{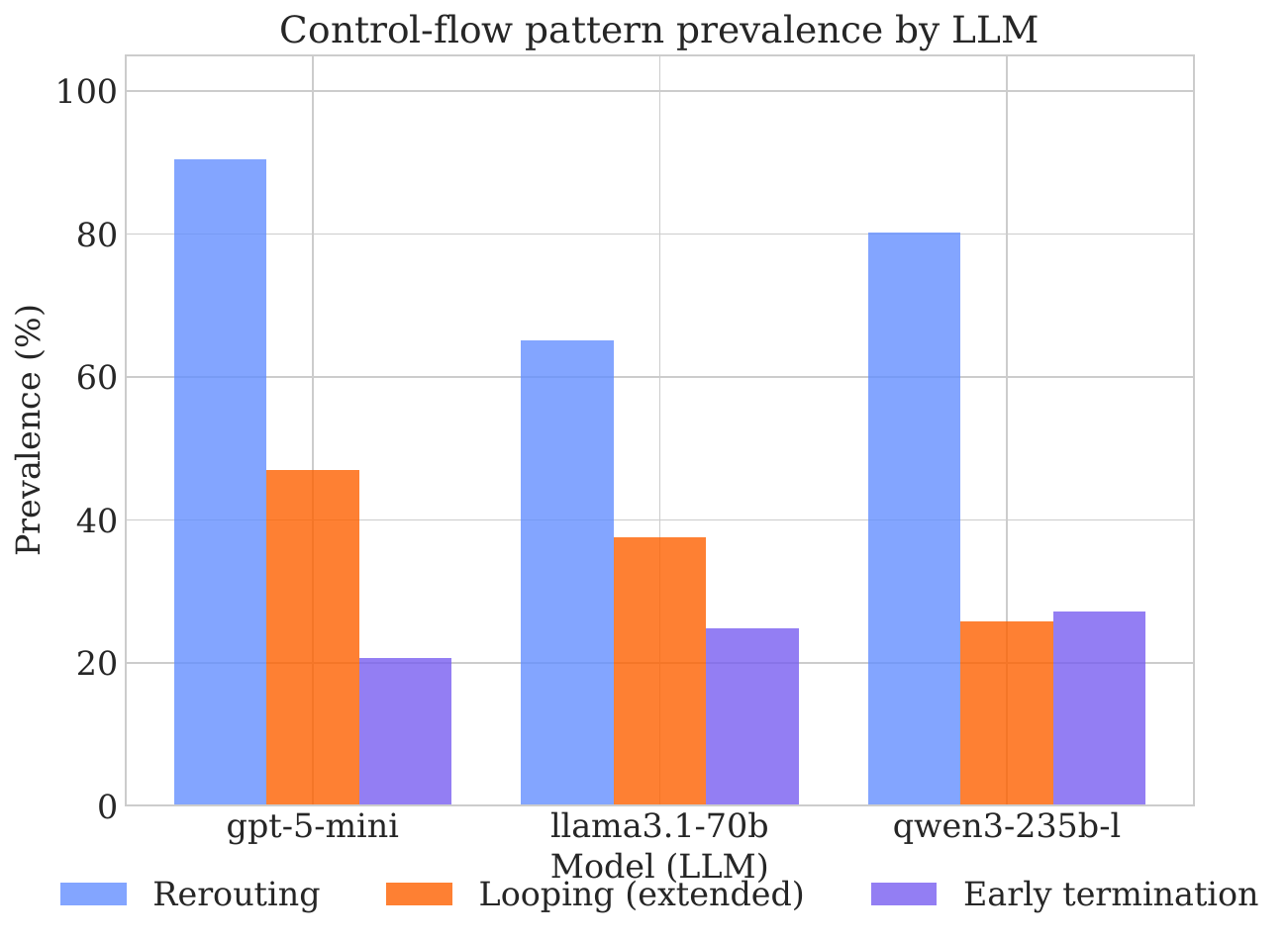}
\caption{Control-flow pattern prevalence by LLM backend.}
\label{fig:llm_control_flow_prevalence}
\end{figure}

\section{Operational Cost and LLMs Comparison}
\label{sec:perturbation_effects}
\label{subsec:operational_cost}

We next map perturbation families to
quantify operational costs induced
by recovery attempts. 
We measure token overhead 
(perturbed tokens / baseline tokens) and examine cost-correctness tradeoffs.

\paragraph{Cost by manifestation type.}
Silent semantic corruption typically incurs near-baseline cost (the workflow ``does not notice''
the semantic drift). Recovery detours are costlier due to retries and additional validation steps.
Divergent failures are bimodal: early termination reduces cost but fails fast, while looping
failures can be extremely expensive.
Figure~\ref{fig:token_overhead} reports token overhead distributions.

\begin{itemize}
\item \textbf{Silent semantic corruption:} Baseline cost (median 1.0$\times$). Execution 
follows nominal path despite corrupted semantics.
\item \textbf{Behavioral detours with recovery:} Substantial overhead (median \costBehavioralDetours, 
IQR 1.08--2.49$\times$). Additional routing, retries, and validation consume disproportionate resources.
\item \textbf{Combined disruption:} Variable cost. Early termination reduces cost (median 0.71$\times$); 
extended execution increases cost (median \costExtendedExecution).
\end{itemize}

\begin{table}[t]
\centering
\caption{Top-5 perturbations (median overhead, gpt-5-mini).}
\vspace{-3pt}
\label{tab:high_cost_perts}
\small
\begin{tabular}{lcc}
\toprule
\textbf{Perturbation} & \textbf{Median Overhead} & \textbf{Recovery Rate}$^*$ \\
\midrule
encoding\_error & 2.4$\times$ & 23.3\% \\
watermark & 2.1$\times$ & 7.0\% \\
text\_redaction & 1.9$\times$ & 17.8\% \\
contrast\_reduction & 1.8$\times$ & 9.3\% \\
ocr\_noise & 1.4$\times$ & 21.7\% \\
\bottomrule
\end{tabular}
\vspace{2pt}\\
\small$^*$ Percentage of perturbed runs where task outcome matches clean baseline.
\end{table}

\vspace{-0.05in}
\textbf{Cost-correctness tradeoff.}
High cost does not guarantee recovery, and low cost does not guarantee correctness.
Workflows face a tradeoff: low-cost executions may miss contamination (silent corruption), while 
high-cost executions sometimes recover correctness. Only \pctHighCostCorrect of high-cost runs (overhead > 2$\times$) 
produce correct answers, while \pctLowCostIncorrect of low-cost runs (overhead < 1.2$\times$) produce incorrect 
answers.

\finding{Cost is a poor indicator of correctness. \pctLowCostIncorrect of low-cost runs produce incorrect answers; 
only \pctHighCostCorrect of high-cost runs succeed. Silent semantic corruption disguises errors with baseline costs, 
making cost-based verification fundamentally insufficient.}

\textbf{High-cost perturbations.}
Table~\ref{tab:high_cost_perts} lists the top-5 perturbations by median token overhead.                                                                                                                            
Notably, all five exhibit relatively low recovery rates (7.0--23.3\%),                                                                                                                                             
indicating that high-cost perturbations tend to be difficult to overcome.                                                                                                                                          
Among these, \texttt{encoding\_error} and \texttt{ocr\_noise} achieve                                                                                                                                              
the highest recovery rates (23.3\% and 21.7\%, respectively).                                                                                                                                                          
In contrast, \texttt{watermark} and \texttt{contrast\_reduction}                                                                                                                                                   
combine high overhead (2.1$\times$ and 1.8$\times$) with                                                                                                                                                         
particularly low recovery rates (7.0\% and 9.3\%),                                                                                                                                                                 
indicating expensive detours that rarely succeed.                                                                                                                                                                  
This disparity highlights the need for targeted mitigations:                                                                                                                                                       
some perturbations warrant retry-based recovery strategies,                                                                                                                                                        
while others may benefit more from early detection and graceful degradation. 

\vspace{-0.01in}
\finding{
High token overhead does not predict recovery. \texttt{encoding\_error} (2.4$\times$) and                                                                                                                
\texttt{ocr\_noise} (1.4$\times$) achieve moderate recovery (23.3\%, 21.7\%), while \texttt{watermark}                                                                                                            
(2.08$\times$) and \texttt{contrast\_reduction} (1.84$\times$) rarely succeed (7.0\%, 9.3\%).                                                                                                                      
Generic cost reduction risks eliminating protective mechanisms.
}

\vspace{-0.15in}
\subsection{LLM Robustness Comparison}
\label{subsec:llm_comparison}

We evaluate robustness across three LLM backends: GPT-5-mini, LLaMA-3.1-70B, and Qwen3-235B. 
Figure~\ref{fig:llm_control_flow_prevalence} compares how different backends express contamination in control flow, 
while Figure~\ref{fig:divergence_step_by_llm} shows first divergence timing distributions: LLaMA-3.1-70B exhibits 
earlier divergence than GPT-5-mini and Qwen3-235B, suggesting faster detection but potentially less 
robustness to initial perturbations.
Differences here indicate that "agent robustness" is partly a property of the model's decision behavior 
(e.g., willingness to retry or re-route), not only the perturbation severity.
This supports reporting model-level robustness in terms of behavioral fingerprints 
(control-flow responses), not just aggregate accuracy.

\finding{LLM backend significantly shapes contamination response. GPT-5-mini exhibits 48.6\% behavioral 
detours with recovery; LLaMA-3.1-70B only 35.4\%; Qwen3-235B 38.3\%. Same perturbations trigger 
different strategies, making model choice a robustness lever.}

\vspace{-0.1in}
\section{Discussion and Future Directions}
\label{sec:discussion}

\subsection{Gaps in Current Guardrails}
\label{subsec:guardrail_gaps}

Contemporary multi-agent frameworks~\cite{langgraph, wu2024autogen, crewai2024} employ guardrails that primarily 
monitor execution health: detecting tool failures, enforcing retry budgets, tracking confidence scores, 
and limiting computational costs~\cite{guardrails_ai, nemo_guardrails2023}. These mechanisms assume that 
contamination manifests as observable control-flow disruption—agents entering error states, tools 
returning malformed outputs, or workflows exceeding resource limits.

Our findings challenge this assumption. Silent semantic corruption (\pctSilentCorruption of runs) preserves nominal 
execution structure while producing incorrect outputs, evading guardrails that trigger on structural 
anomalies. Conversely, behavioral detours (\pctBehavioralDetours) exhibit substantial structural divergence yet recover 
correct answers, potentially triggering false alarms in systems optimized for execution stability. 
Cost-based limits risk eliminating protective retry mechanisms while 
tolerating low-cost silent failures. Current verification practices optimize for preventing runaway 
costs and control-flow failures, missing the semantic correctness and execution brittleness that dominate 
contamination in artifact-driven workflows. For practitioners, this suggests that divergence should 
not be treated as a standalone failure signal, but interpreted jointly with timing, 
control-flow pattern, modality, and deployment objective. 
Early divergence is more suggestive of foundational extraction failures, 
while later divergence points to reasoning-stage sensitivity; 
rerouting, looping, and early termination imply different interventions. 
The appropriate response is therefore use-case dependent: 
low-latency systems may use divergence primarily for triage, 
whereas high-stakes settings may justify more aggressive validation despite additional cost.

\vspace{-3pt}
\subsection{Scope and Generalization}
\label{subsec:scope_validity}

Our experiments intentionally fix the orchestration apparatus to isolate perturbation effects 
from orchestration drift. Accordingly, exact prevalence rates for rerouting, 
looping, early termination, and recovery should not be interpreted as architecture-invariant quantities. 
Different orchestration strategies, validation interfaces, and recovery mechanisms 
can alter whether corruption is corrected or amplified. We already observe such variation across the three backends 
studied here. At the same time, the central qualitative finding remains stable across all three. 
We therefore view the exact rates as setup-dependent, 
while treating relative patterns across conditions as the primary object of interpretation.

The \numTasks GAIA tasks were selected because they require heterogeneous artifact processing 
and multi-step coordination, which are the properties needed to study contamination propagation. 
Our goal is therefore not to claim exhaustive coverage of multi-agent workloads, 
but to establish the phenomenon under controlled conditions on tasks where 
externally derived information materially shapes downstream decisions. 
Longer-horizon workflows are an important extension. 
As execution traces grow, contamination may have more opportunities both to accumulate 
and to be corrected downstream, so quantitative rates may shift with task horizon 
and orchestration style. However, we expect the broader qualitative conclusion 
to remain unchanged.

\vspace{-3pt}
\subsection{Open Research Directions}
\label{subsec:future_directions}

\paragraph{Contamination-origin attribution and causal tracing.}
Our first divergence point $t^\star$ localizes \emph{when} contamination manifests, but not \emph{which} 
upstream extraction or transformation introduced it. A natural next step is provenance-based origin 
attribution: backtracking from the first divergent event to rank candidate upstream artifacts by likely 
influence on downstream decisions. Since our logs already encode dependency links across tool results, 
memory updates, and agent messages, this analysis can be layered on top of the current instrumentation 
without changing the workflow architecture. Candidate origins can then be stress-tested with targeted 
replay or ablation to separate correlation from causation. An important open challenge is 
attribution uncertainty when multiple correlated artifacts co-occur; confidence calibration for root-cause 
claims will therefore be as important as raw localization accuracy. This would move analysis from temporal 
localization to actionable origin attribution (e.g., OCR extraction error, schema misalignment during 
transformation, or downstream reasoning-stage misuse).
\vspace{-4pt}
\paragraph{Learning contamination-resilient workflows.}
Given modality-specific manifestation patterns (tabular favoring extended execution, audio favoring 
early termination), can workflow architectures be learned or adapted to minimize contamination 
propagation? Reinforcement learning approaches could optimize routing policies for robustness under 
perturbation, or meta-learning could identify which agent specializations reduce cross-boundary 
contamination. The cost-correctness decoupling suggests that standard accuracy-maximizing objectives 
are insufficient—multi-objective optimization balancing outcome correctness, structural stability, 
and operational cost may be necessary.
\vspace{-4pt}
\paragraph{Adaptive verification budgets and risk-proportional validation.}
Static guardrails apply uniform verification regardless of evidence quality or task criticality. 
Our findings suggest stratified verification: high-confidence extractions warrant aggressive retries; 
low-confidence inputs should trigger early validation or human-in-the-loop checkpoints; high-stakes 
domains (clinical, financial) demand comprehensive trace auditing. Can machine learning meta-models 
predict contamination likelihood from extraction features (cross-modality consistency, parsing confidence, 
historical failure rates) and dynamically allocate verification resources? What are the fundamental 
tradeoffs between verification cost and contamination detection coverage?
\vspace{-4pt}
\paragraph{Trace-native evaluation and benchmark design.}
Current benchmarks measure endpoint accuracy, collapsing internal workflow dynamics. Our work demonstrates 
that outcome-only metrics miss \pctBehavioralDetours of runs with substantial structural divergence. Future benchmarks 
should expose execution traces alongside answers, enabling robustness evaluation along both dimensions. 
What trace divergence thresholds indicate fragile vs. adaptive behavior? How should benchmark datasets 
be constructed to cover diverse manifestation patterns rather than focusing solely on task difficulty?
\vspace{-4pt}
\paragraph{Cross-domain generalization of contamination patterns.}
Our study focuses on GAIA tasks with specific modalities. Do manifestation patterns generalize to other 
domains (legal document analysis, scientific literature review, code generation from specifications)? 
Are there domain-specific failure modes not captured by our taxonomy? Investigating contamination in 
long-horizon workflows (multi-day research synthesis, iterative debugging) may reveal new challenges 
around contamination accumulation and compounding across extended interaction sequences.

\vspace{-3pt}
\section{Conclusion}
\label{sec:conclusion}

Multi-agent workflows must be resilient to corrupted externally-derived information 
that appears locally plausible yet distorts downstream computation.
We conducted trace-level analysis of \numRuns runs across \numTasks GAIA tasks and \numLLMs language models
to understand how contamination propagates and manifests.
We find that structural divergence and outcome correctness are decoupled: 
workflows diverge substantially yet recover (\pctBehavioralDetours), or remain stable yet fail (\pctSilentCorruption), 
with distinct control-flow signatures and modality-specific costs that guide targeted defense.

\section*{Acknowledgments}

We thank the anonymous reviewers for their constructive feedback. 
This research was supported by a gift to the LinkedIn–Cornell 
Bowers Strategic Partnership, ARO grant W911NF-25-1-0254, 
BSF grant 2024101 and a grant from Infosys.

\bibliographystyle{ACM-Reference-Format}
\bibliography{ref}

\clearpage

\appendix
\section{Implementation Details}
\label{sec:appendix_impl}

\paragraph{Agent roles.}
We use the following role set (a subset may be inactive on tasks that do not require the
corresponding modality):
\begin{itemize}
  \item \textbf{Data Analyst}: parses and analyzes CSV/XLSX tables; performs aggregations and joins.
  \item \textbf{Document Analyst}: extracts and summarizes content from PDFs/DOCX/PPTX.
  \item \textbf{Visual Analyst}: interprets images when present.
  \item \textbf{Audio Analyst}: transcribes/analyzes audio when present.
  \item \textbf{Computation Agent}: executes programmatic computations and consistency checks.
  \item \textbf{Fact Checker}: cross-checks claims against cited evidence/provenance within the attachment bundle.
  \item \textbf{Synthesizer}: aggregates intermediate outputs into the task outcome with citations.
\end{itemize}

\paragraph{Agent tooling.}
We deploy a specialized agent toolkit, where each agent is equipped with specific tools and libraries 
to handle different modalities and computational tasks. Table~\ref{tab:agent_tools} summarizes the agents, 
their tools, and the Python libraries they utilize.

\begin{table}[H]
\centering
\caption{Modality-specific tools and associated Python libraries.}
\label{tab:agent_tools}
\footnotesize
\setlength{\tabcolsep}{4pt}
\renewcommand{\arraystretch}{1.0}
\begin{tabularx}{\columnwidth}{X X}
\toprule
\textbf{Tools} & \textbf{Python Libraries} \\
\midrule
Excel analysis, Python execution 
    & pandas, openpyxl \\

Python code execution 
    & pandas, numpy \\

Math evaluation 
    & math (stdlib) \\

Image analysis 
    & PIL/Pillow, base64, Vision LLM \\

PDF, PPTX, DOCX parsing 
    & PyMuPDF, pdfplumber, python-pptx, python-docx \\

Web search \& fetch 
    & requests \\

Audio transcription 
    & openai (Whisper API), mutagen \\
\bottomrule
\end{tabularx}
\end{table}

\subsection{Controlled experimental setup}
\label{subsec:appendix_controlled}

To isolate the effect of perturbations, we hold the workflow configuration fixed across 
clean and perturbed runs. Specifically, we fix:
\begin{itemize}
  \item \textbf{Agent roles and policies}: Each agent's role, system prompt, and decision logic remain unchanged.
  \item \textbf{Tool wrappers and libraries}: All artifact-processing tool implementations and versions are identical.
  \item \textbf{Shared-state schema}: The structure and access patterns for the shared memory workspace remain constant.
  \item \textbf{Stopping and retry policies}: Conditions for terminating execution or triggering retries are held fixed.
  \item \textbf{Random seeds}: When applicable, all stochastic processes use identical seeds to eliminate sampling variance.
\end{itemize}
This enables paired comparisons of clean and perturbed traces that isolate the effect 
of information corruption from orchestration drift or policy adaptation.

\subsection{Workflow apparatus architecture}
\label{subsec:appendix_architecture}

Our experiments instantiate structured workflows as a multi-agent orchestration with a shared 
workspace, treated as experimental apparatus rather than a proposed production architecture. 
This design choice prioritizes comparability and traceability over generality or adaptability.


\subsubsection{Execution flow and shared state}
A coordinator agent examines the current shared memory and routing metadata 
(e.g., task type, required modalities, completion status) to select the next specialized agent. 
Agents communicate indirectly via shared memory: each agent reads relevant memory entries, 
performs its work, and writes typed results (e.g., extracted tables, computed values, synthesis notes) 
back to shared memory. This decoupled architecture enables clean instrumentation of information flow 
and supports the provenance tracking required for contamination analysis.

\subsection{Perturbation injection model}
\label{subsec:appendix_perturbation}

For a selected upstream information item $x$ (e.g., an extracted table cell value 
or a parsed text span), we apply a perturbation operator $\pi$ 
to obtain $\tilde{x} = \pi(x)$, yielding a perturbed run trace $\tilde{\tau}$.

\textbf{Injection locus.} We inject perturbations primarily at the level of 
artifact-derived representations (the objects consumed by downstream agents), 
rather than modifying raw source files directly. This reflects realistic failure modes: 
extraction/parsing errors and transduced representation errors 
are more common than corrupted source files.

\textbf{Reproducibility.} Perturbations are generated using fixed random seeds, 
and we record for each run: perturbation type, injection locus, affected evidence identifiers, 
and any relevant parameters (e.g., noise magnitude, mutation target).

\subsection{Perturbation types and rationale}
\label{subsec:appendix_perturbation_types_rationale}
We apply a range of perturbation types that reflect realistic failure modes across modalities,
targeting content corruption, structure/format corruption, provenance corruption, and tool reliability noise.
These perturbations are designed to be locally plausible (e.g., a misaligned table parse still
produces a valid table structure) to test the workflow's ability to detect and contain corrupted evidence.

\begin{table}[H]
\centering
\caption{Modality-specific perturbation operators. Perturbations operationalize the uncertainty classes.}
\label{tab:perturbations}
\setlength{\tabcolsep}{6pt}
\renewcommand{\arraystretch}{1.35} 

\begin{tabularx}{\linewidth}{@{} L{0.34\linewidth} L{0.62\linewidth} @{}}
\toprule
\textbf{File Type} & \textbf{Perturbations} \\
\midrule
Tabular (CSV/XLSX/JSON) &
\texttt{column\_swap}, \texttt{label\_corrupt}, \texttt{data\_type\_corrupt},\\[-4pt]
& \texttt{row\_duplicate}, \texttt{irrelevant\_columns}, \texttt{unit\_change} \\[5pt]
Documents (PDF/TXT/DOCX/PPTX) &
\texttt{ocr\_noise}, \texttt{number\_corruption}, \texttt{text\_redaction},\\[-4pt]
& \texttt{paragraph\_shuffle}, \texttt{encoding\_error}, \texttt{section\_removal} \\[5pt]
Images (PNG/JPG) &
\texttt{blur}, \texttt{noise}, \texttt{low\_resolution}, \texttt{partial\_occlusion},\\[-4pt]
& \texttt{contrast\_reduction}, \texttt{watermark} \\[4pt]
Audio (MP3/WAV) &
\texttt{background\_noise}, \texttt{speed\_change}, \texttt{low\_pass\_filter} \\[5pt]
\bottomrule
\end{tabularx}
\end{table}

\subsubsection{Tabular} Tabular artifacts (CSV/XLSX) are vulnerable to both content and structure corruption. 
A misaligned parse can shift entire rows or columns, causing downstream queries to reference wrong data. 
Other noise types include header confusion, numeric noise, unit mismatches, and provenance drift
(e.g., citing the wrong cell). The tabular perturbations used in our experiments 
are summarized in Table~\ref{tab:tabular_perturbations}.

\subsubsection{Documents} To mimic common document (PDF/DOCX/PPTX) extraction errors, we apply perturbations that simulate misread spans, 
missing qualifiers, and layout errors. For example, an OCR misread can corrupt a critical numeric constraint,
while a layout error can reorder paragraphs and shift context.
The document perturbations used in our experiments are summarized in Table~\ref{tab:document_perturbations}.

\begin{table}[H]
\centering
\caption{Tabular perturbations (applied to parsed table representations).}
\label{tab:tabular_perturbations}
\small
\begin{tabular}{p{0.24\linewidth} p{0.68\linewidth}}
\toprule
\textbf{Perturbation} & \textbf{Description} \\
\midrule
Row/column misalignment & Shifts a contiguous block of cells by $\pm 1$ row/column (structure corruption). \\
Header drift / confusion & Swaps header labels or promotes a footnote row into the header region (structure corruption). \\
Numeric perturbation & Adds multiplicative noise to selected numeric cells (content corruption; intensity controls \#cells). \\
Unit mismatch & Applies a unit conversion to values without updating the label (content+provenance ambiguity). \\
Cell reference drift & Corrupts provenance metadata (e.g., cites B12 when value came from B13) (provenance corruption). \\
\bottomrule
\end{tabular}
\end{table}

\begin{table}[H]
\centering
\caption{Document perturbations (applied to extracted spans / structured representations).}
\label{tab:document_perturbations}
\small
\begin{tabular}{p{0.24\linewidth} p{0.68\linewidth}}
\toprule
\textbf{Perturbation} & \textbf{Description} \\
\midrule
Omission of critical span & Removes a task-critical sentence or qualifier (e.g., ``excluding tax'') (content corruption). \\
Insertion of plausible snippet & Inserts a plausible but false constraint/value near the relevant span (content corruption). \\
Ordering / layout error & Reorders a small set of paragraphs or simulates column-order swaps (structure corruption). \\
Numeric/date corruption & Perturbs key numbers/dates by a controlled factor (content corruption). \\
Citation pointer shift & Keeps text unchanged but shifts page/offset provenance by $\pm 1$ (provenance corruption). \\
Tool truncation & Simulates partial extraction (e.g., truncated output length) (tool reliability noise). \\
\bottomrule
\end{tabular}
\end{table}

\subsubsection{Images and audio}
\label{subsubsec:img_audio}
For tasks with image/audio attachments, we apply perturbations that primarily stress extraction
reliability (OCR/ASR brittleness) and partial observability. Because our core focus is workflow
propagation rather than perceptual robustness, we restrict to a small set of lightweight,
interpretable corruptions and report these results separately when sample sizes are sufficient.
\begin{itemize}
  \item \textbf{Images:} partial occlusion of a task-critical region; downscale/upscale to induce OCR errors.
  \item \textbf{Audio:} additive background noise at a fixed SNR; mild time-scale modification.
\end{itemize}

\section{Trace Event Schema}
\label{sec:appendix_events}

For reference, we provide the complete structured event schema used in trace logging.
See Table~\ref{tab:trace_events} in the main text for the core event types and fields 
used in divergence analysis. Additional event types available for detailed post-hoc analysis include:
\begin{itemize}
  \item \texttt{memory\_read}: Agent reads from shared memory (logged for provenance tracking).
  \item \texttt{retrieval\_shown}: Retrieval results displayed to agent (when applicable).
  \item \texttt{tool\_failure}: Tool execution failed or timed out.
  \item \texttt{agent\_halt}: Execution stopped (early termination or max steps reached).
\end{itemize}

\begin{table}[t]
\centering
\small
\begin{tabular}{
    >{\raggedright\arraybackslash}p{0.26\linewidth} 
    >{\raggedright\arraybackslash}p{0.29\linewidth} 
    >{\raggedright\arraybackslash}p{0.33\linewidth}}
\toprule
\textbf{Event type} & \textbf{Purpose} & \textbf{Key fields used in divergence analysis} \\
\midrule
\texttt{routing\_decision} & Next-agent selection & \texttt{chosen\_agent} \\
\texttt{tool\_invocation} & External tool call & \texttt{tool\_name}, \texttt{operation}, \texttt{params}, \texttt{success} \\
\texttt{memory\_write} & Shared-state update & \texttt{entry\_id}, \texttt{entry\_type} \\
\texttt{agent\_output} & Agent produces output & \texttt{action}, \texttt{is\_task\_outcome} \\
\texttt{task\_outcome} & Task outcome set & \texttt{answer}  \\
\bottomrule
\end{tabular}
\caption{Structured execution events logged for each run. We use a compact subset of event fields 
for structural divergence analysis and ignore lexical content, IDs, and timestamps.}
\label{tab:trace_events}
\end{table}

\clearpage

\section{LLM-Specific Results: LLaMA and Qwen}
\label{sec:appendix_llm_results}

This appendix provides detailed analysis for LLaMA-3.1-70B and Qwen3-235B backends, using the same 
figures as the main paper (GPT-5-mini) to enable direct comparison.

\subsection{LLaMA-3.1-70B Results}

\begin{figure}[b]
\centering
\includegraphics[width=1.0\columnwidth]{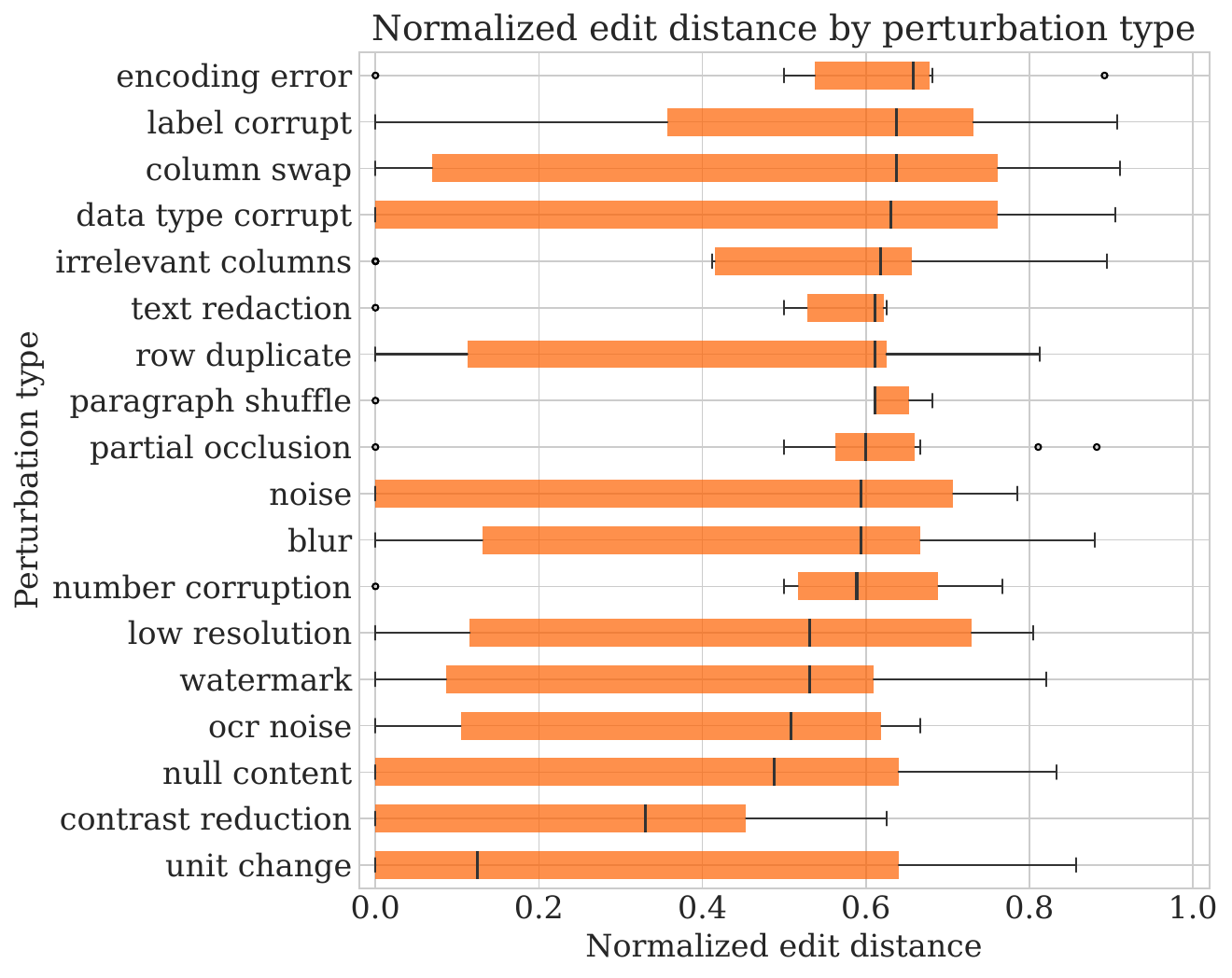}
\caption{Trace divergence (normalized edit distance) by perturbation type --- LLaMA-3.1-70B.}
\label{fig:llama_edit_distance}
\end{figure}

\begin{figure}[b]
\centering
\includegraphics[width=1.0\columnwidth]{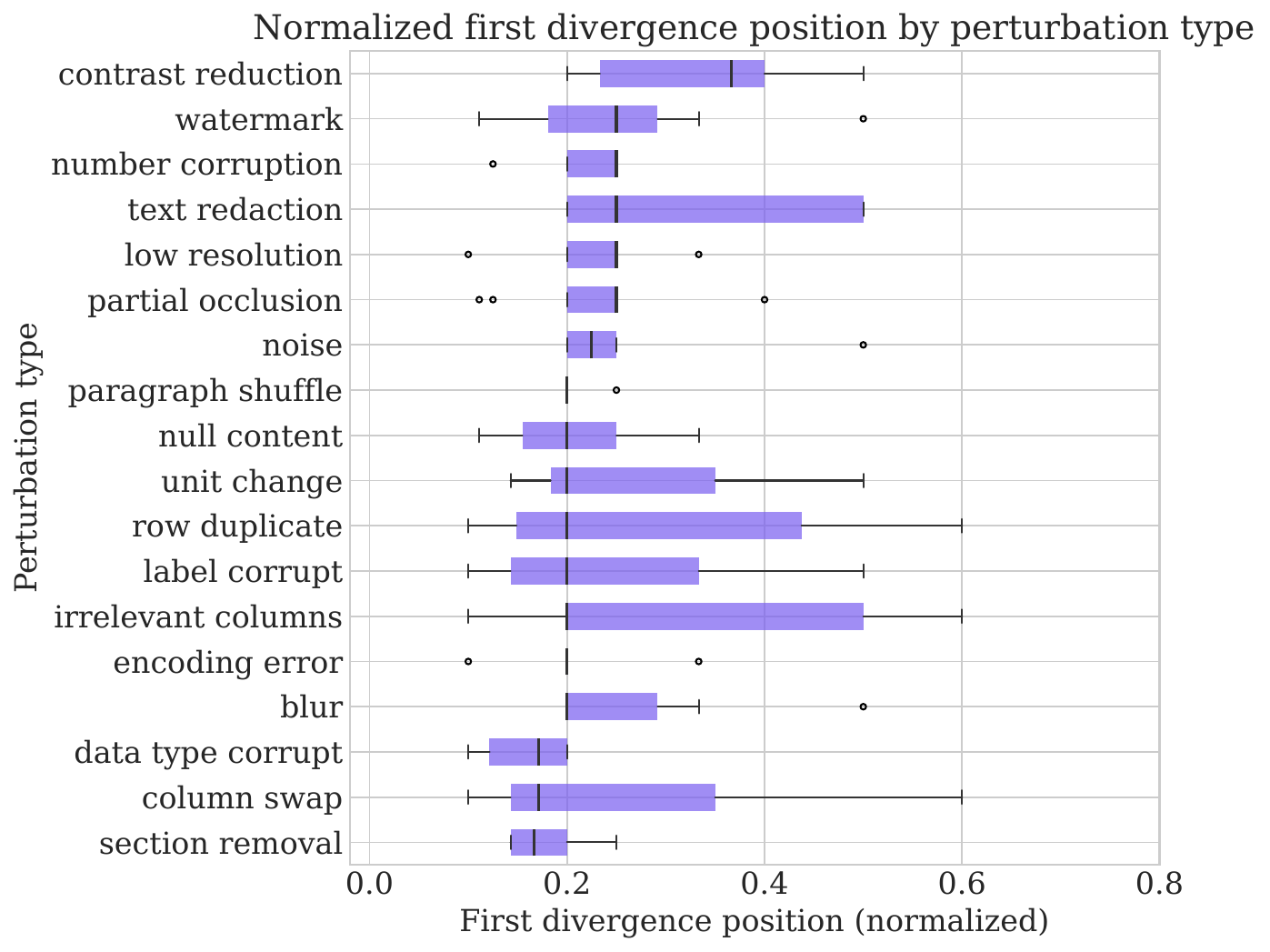}
\caption{First divergence point timing by perturbation type --- LLaMA-3.1-70B.}
\label{fig:llama_divergence_timing}
\end{figure}

\begin{figure}[b]
\centering
\includegraphics[width=1.0\columnwidth]{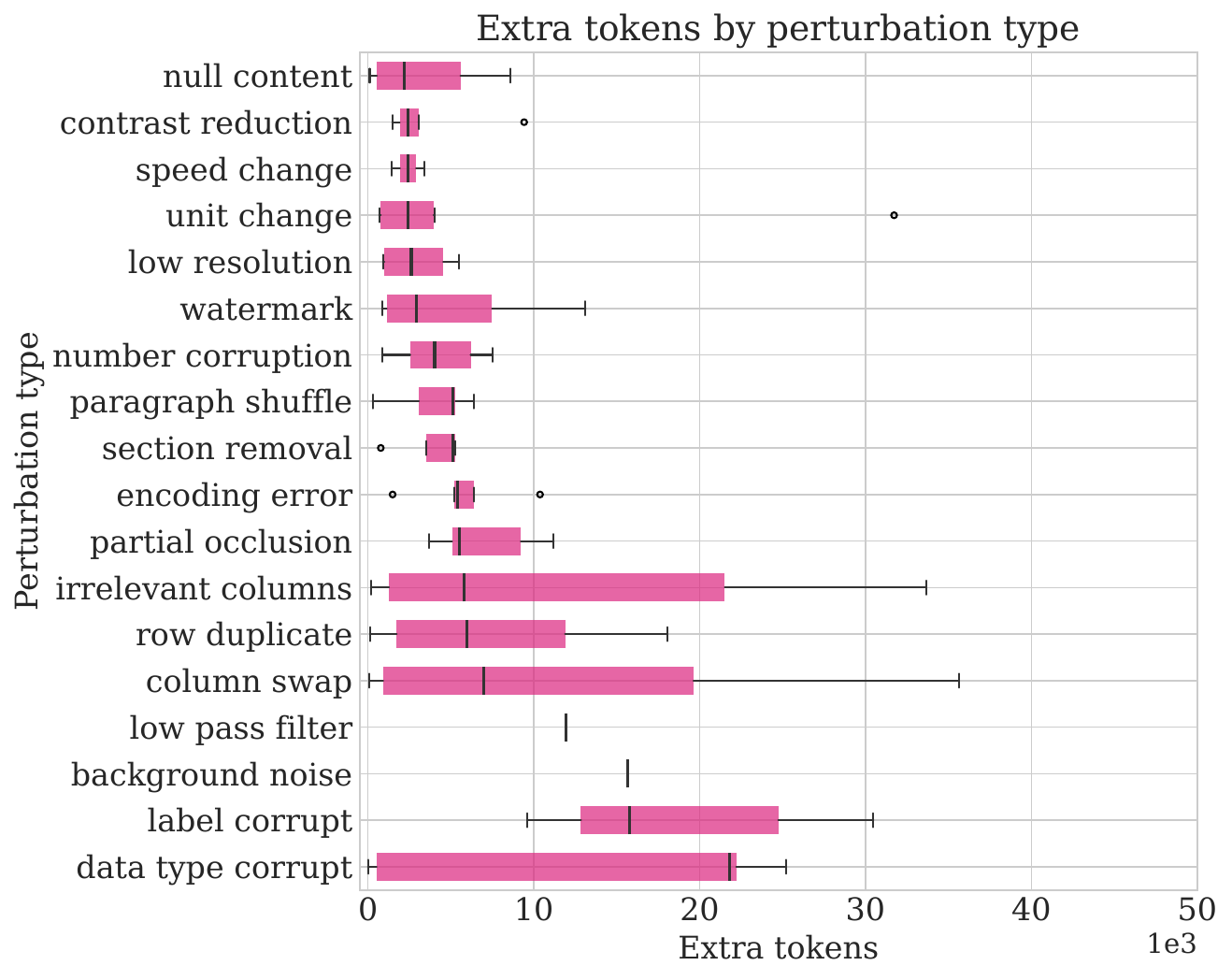}
\caption{Token overhead by perturbation type --- LLaMA-3.1-70B.}
\label{fig:llama_token_overhead}
\end{figure}

\begin{figure}[b]
\centering
\includegraphics[width=1.0\columnwidth]{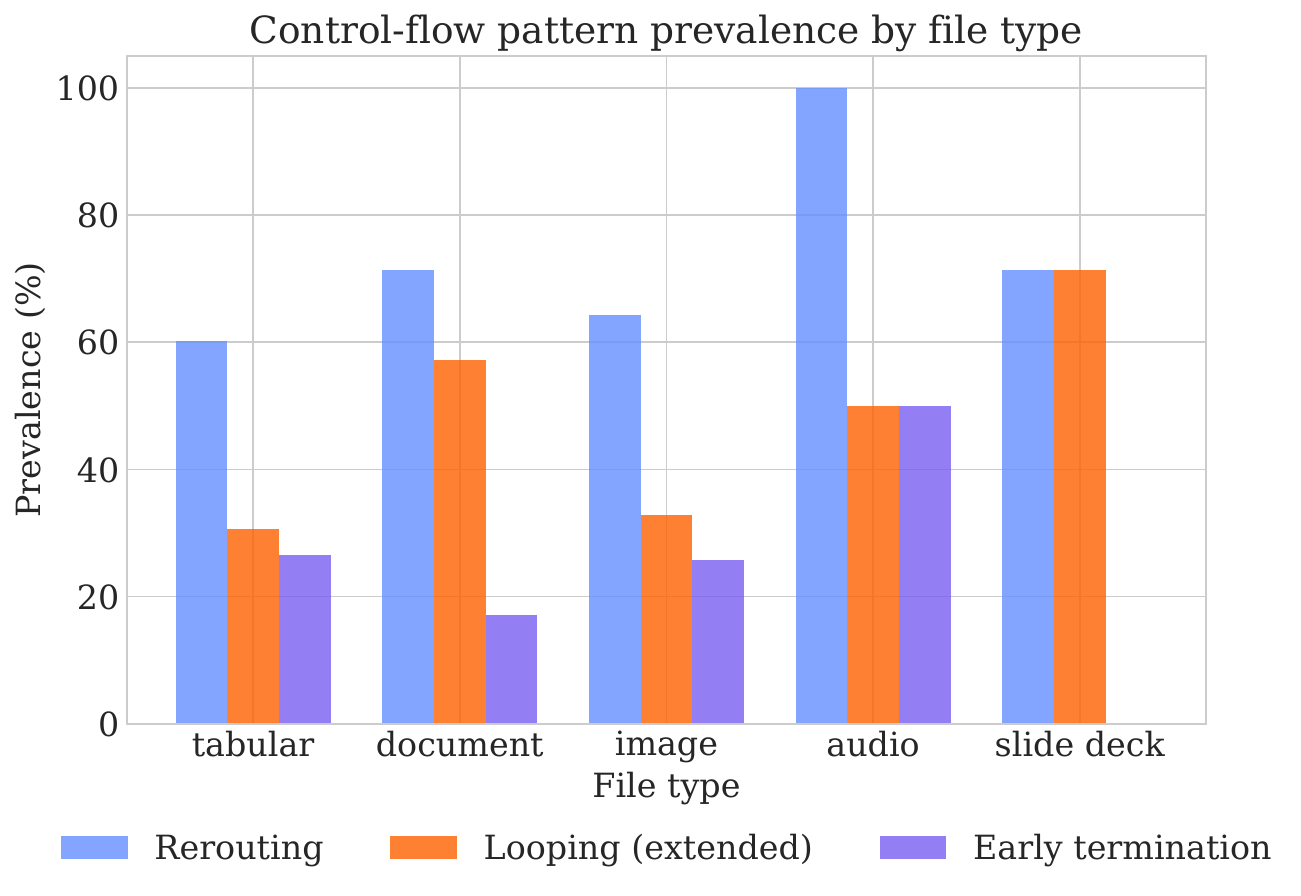}
\caption{Control-flow patterns (rerouting, looping, termination) by artifact modality --- LLaMA-3.1-70B.}
\label{fig:llama_control_flow_by_file_type}
\end{figure}

\begin{figure}[b]
\centering
\includegraphics[width=1.0\columnwidth]{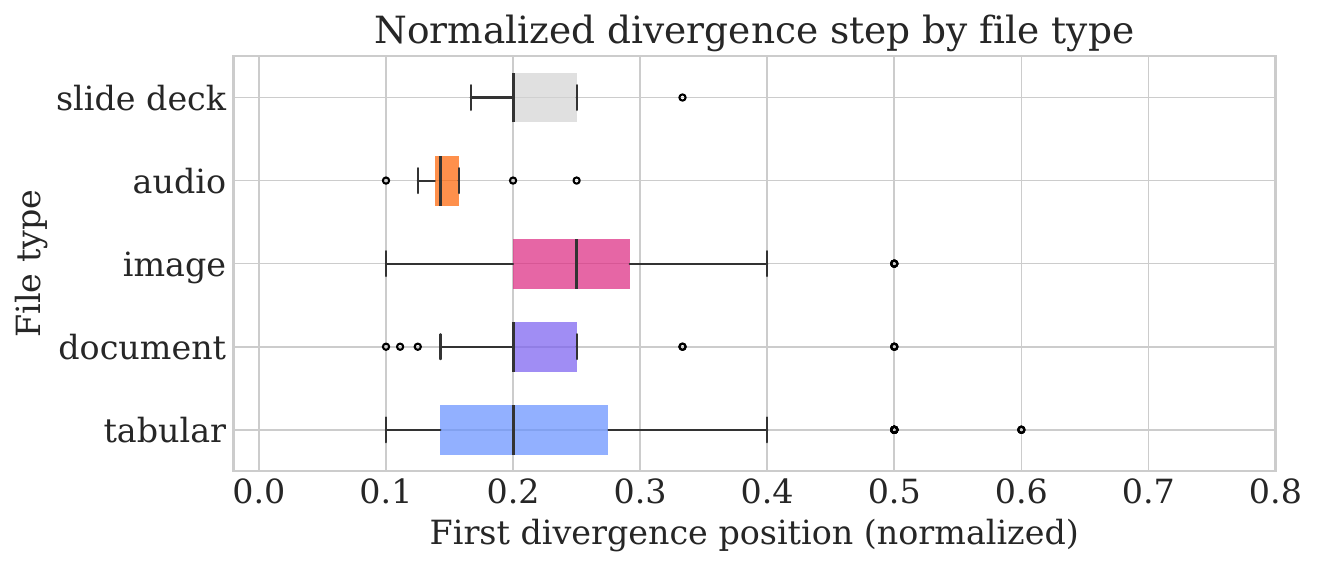}
\caption{First divergence point timing by artifact modality --- LLaMA-3.1-70B.}
\label{fig:llama_divergence_step_by_file_type}
\end{figure}

LLaMA-3.1-70B exhibits distinct contamination response patterns compared to GPT-5-mini. 
Figure~\ref{fig:llama_edit_distance} shows trace divergence by perturbation type, while 
Figure~\ref{fig:llama_divergence_timing} reveals when divergence manifests temporally. 
Token overhead analysis (Figure~\ref{fig:llama_token_overhead}) indicates how much additional 
computation LLaMA invokes under contamination. Cross-modality analysis (Figures~\ref{fig:llama_control_flow_by_file_type} and ~\ref{fig:llama_divergence_step_by_file_type}) 
demonstrates how artifact type influences control-flow behavior and divergence timing in LLaMA workflows.

\subsection{Qwen3-235B Results}

\begin{figure}[b]
\centering
\includegraphics[width=1.0\columnwidth]{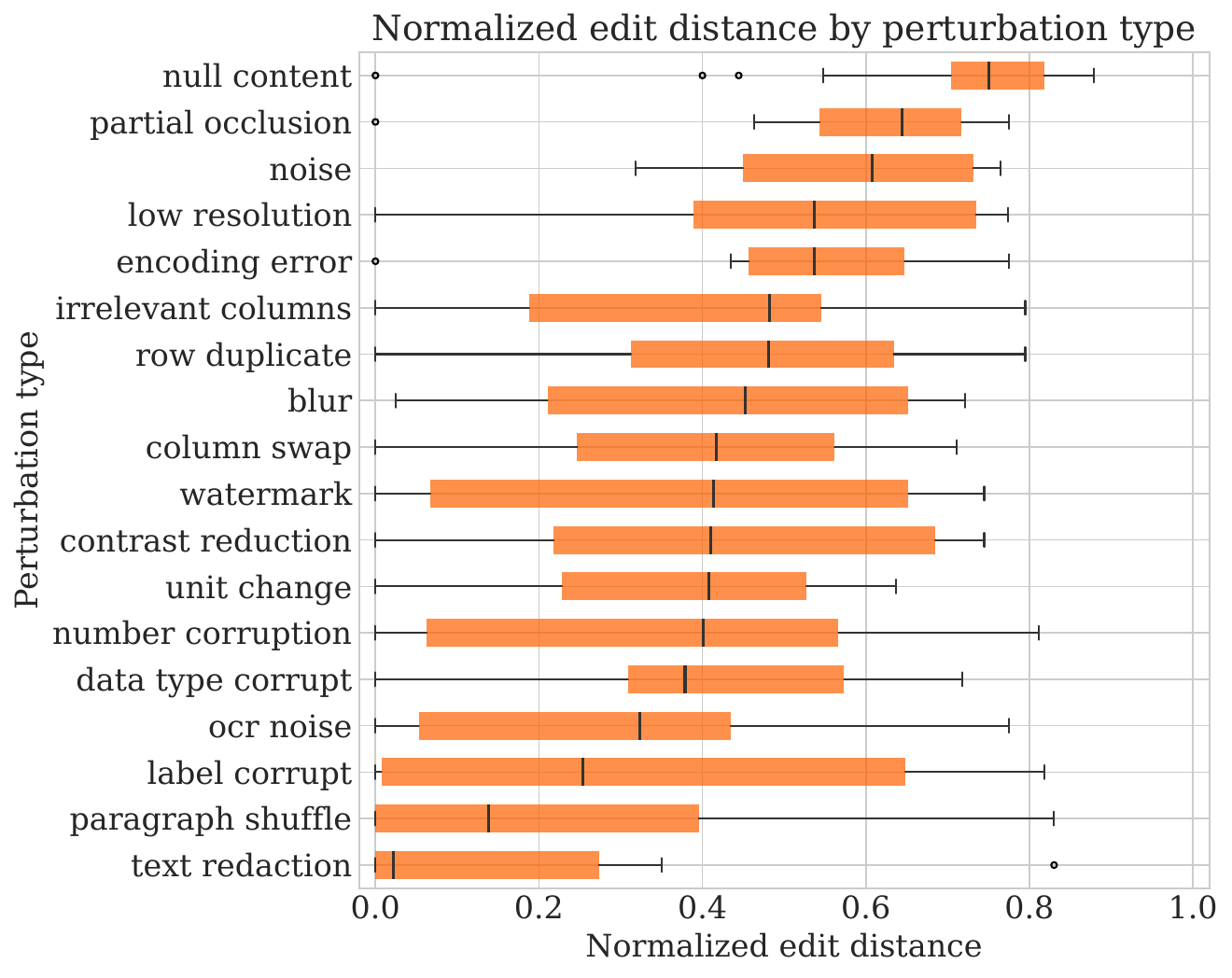}
\caption{Trace divergence (normalized edit distance) by perturbation type --- Qwen3-235B.}
\label{fig:qwen_edit_distance}
\end{figure}

\begin{figure}[b]
\centering
\includegraphics[width=1.0\columnwidth]{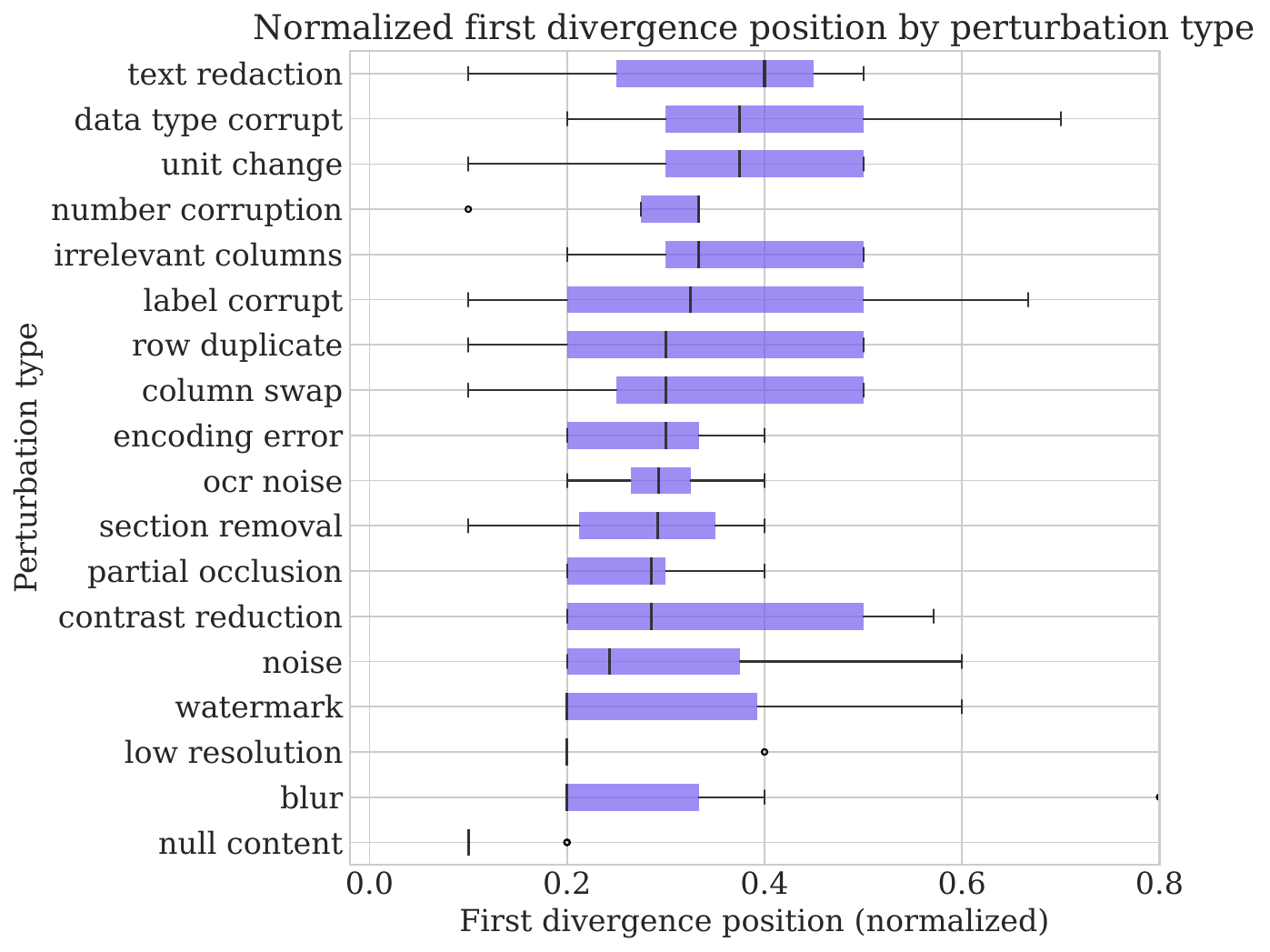}
\caption{First divergence point timing by perturbation type --- Qwen3-235B.}
\label{fig:qwen_divergence_timing}
\end{figure}

\begin{figure}[b]
\centering
\includegraphics[width=1.0\columnwidth]{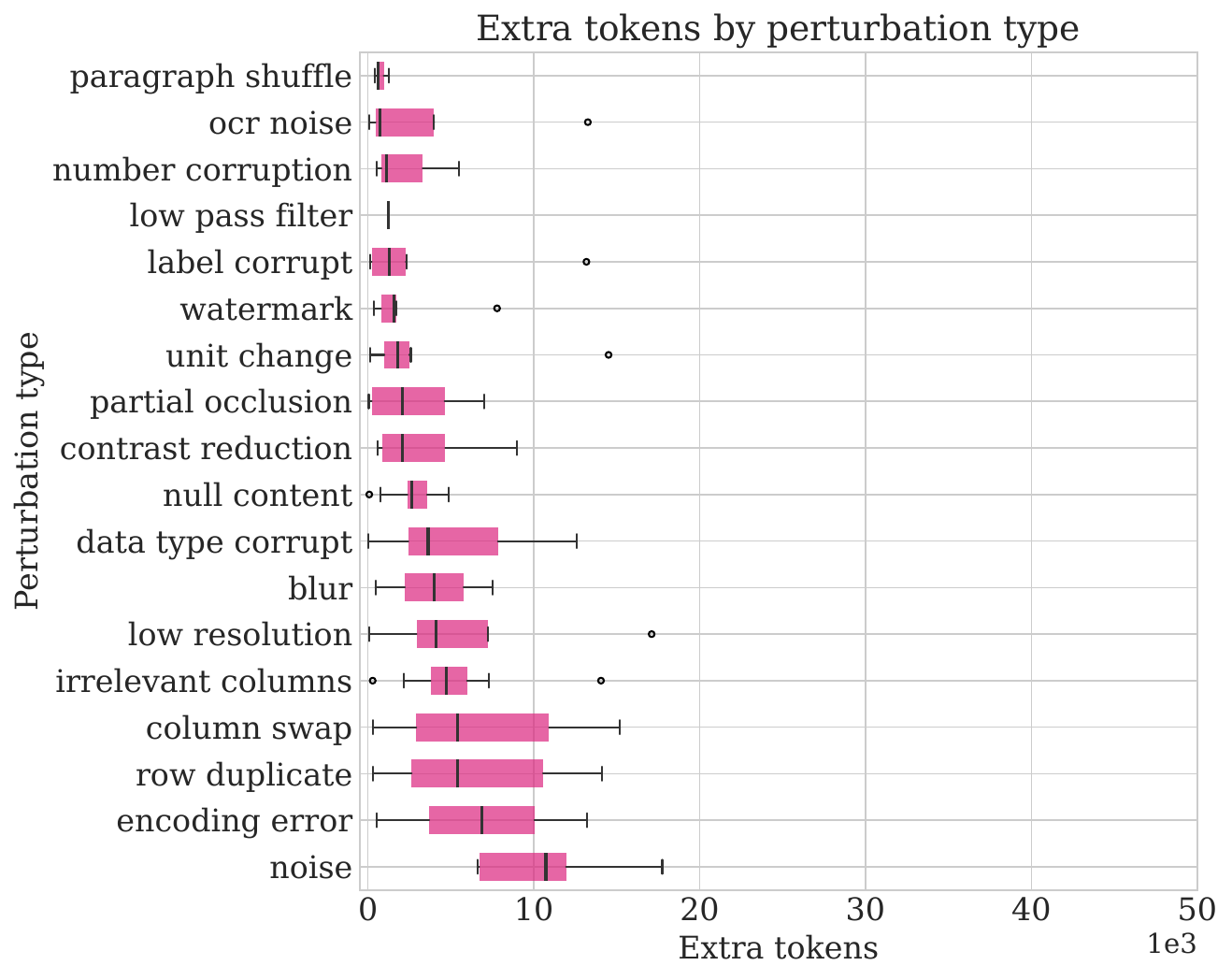}
\caption{Token overhead by perturbation type --- Qwen3-235B.}
\label{fig:qwen_token_overhead}
\end{figure}

\begin{figure}[b]
\centering
\includegraphics[width=1.0\columnwidth]{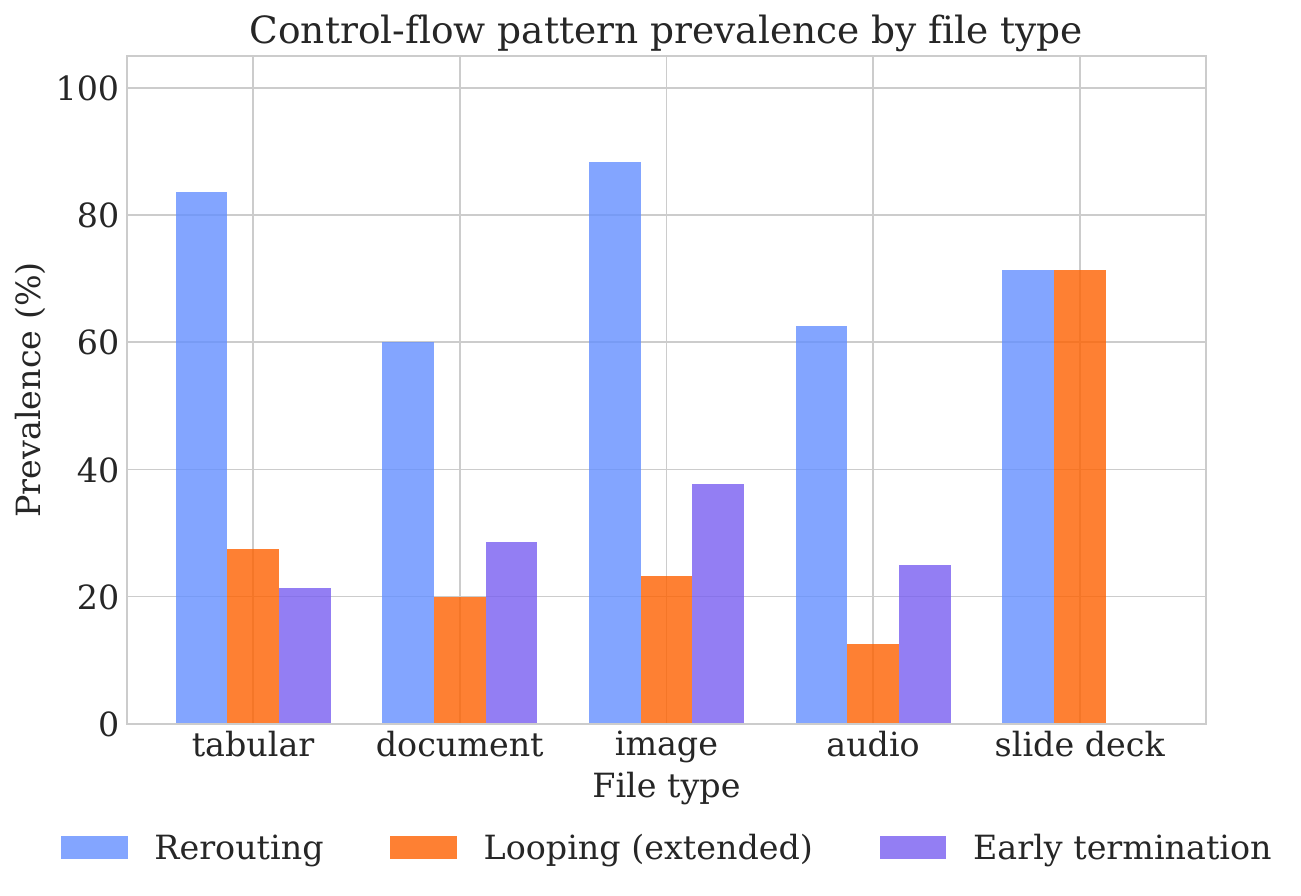}
\caption{Control-flow patterns (rerouting, looping, termination) by artifact modality --- Qwen3-235B.}
\label{fig:qwen_control_flow_by_file_type}
\end{figure}

\begin{figure}[b]
\centering
\includegraphics[width=1.0\columnwidth]{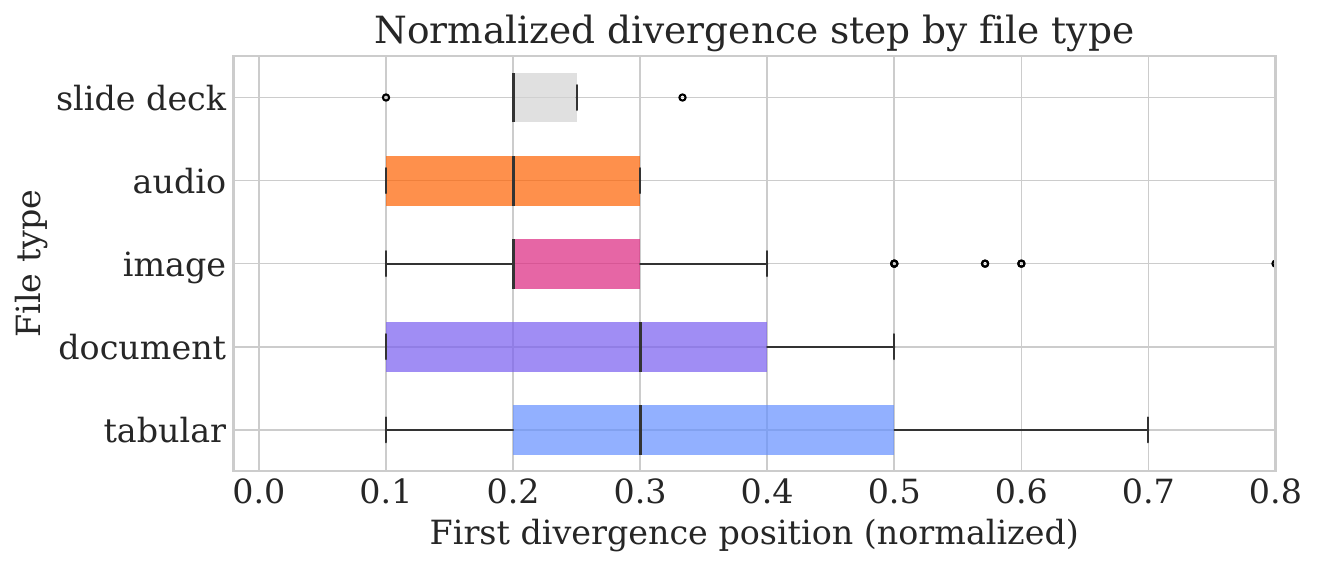}
\caption{First divergence point timing by artifact modality --- Qwen3-235B.}
\label{fig:qwen_divergence_step_by_file_type}
\end{figure}

Qwen3-235B demonstrates yet another robustness profile under contamination. 
Figure~\ref{fig:qwen_edit_distance} presents trace divergence patterns by perturbation type, 
complementing the temporal analysis in Figure~\ref{fig:qwen_divergence_timing}. 
Token overhead comparisons (Figure~\ref{fig:qwen_token_overhead}) show the computational cost Qwen incurs, 
while modality-specific breakdowns (Figures~\ref{fig:qwen_control_flow_by_file_type} and ~\ref{fig:qwen_divergence_step_by_file_type}) 
reveal how different artifact types trigger different control-flow signatures and divergence timing profiles. 
These results enable direct comparison of how model architecture and capability shape contamination resilience.

\end{document}